\documentclass{article}

\PassOptionsToPackage{numbers,sort}{natbib}

\usepackage[preprint]{neurips_2024}

\usepackage[utf8]{inputenc} 
\usepackage[T1]{fontenc}    
\usepackage{url}            
\usepackage{microtype}      
\usepackage{xcolor}         

\usepackage{mathtools}
\usepackage{amsthm}
\usepackage{xspace}
\usepackage{colortbl}
\usepackage{soul}
\definecolor{Gray}{gray}{0.92}
\definecolor{Gray2}{gray}{0.85}
\sethlcolor{Gray}
\definecolor{Newresult}{HTML}{bdf0dd}
\usepackage{lipsum}

\usepackage{enumitem}
\setlist[itemize]{itemsep=2pt}
\setlist[enumerate]{itemsep=2pt}
\usepackage{todonotes}  
\usepackage{nicefrac}
\usepackage{amsmath}
\usepackage{amssymb}
\usepackage{amsfonts}
\usepackage{booktabs}
\usepackage{multicol}
\usepackage{multirow}
\usepackage{blindtext}
\usepackage{floatrow}
\usepackage{caption}
\usepackage{subfig}
\usepackage{wrapfig}
\usepackage{pifont}  
\usepackage{comment}
\usepackage{threeparttable}
\usepackage[hang,flushmargin]{footmisc}
\usepackage{cuted}
\usepackage{epsfig}
\usepackage{graphicx}
\usepackage{outlines}
\usepackage{xfrac}
\usepackage{bm}
\usepackage[retainorgcmds]{IEEEtrantools}  

\usepackage{pifont}  
\usepackage{pdfrender}  
\usepackage{longtable}

\setlist[enumerate]{leftmargin=*}

\usepackage[most]{tcolorbox}
\definecolor{mygreen}{HTML}{39A3B5}
\definecolor{darkmygreen}{HTML}{2596be}
\tcbset{on line, 
        boxsep=4pt,left=-1pt,right=-1pt,top=-2.6pt,bottom=-2.6pt,
        colframe=mygreen,colback=mygreen,boxrule=0pt,
        highlight math style={enhanced}
        }

\definecolor{tumblue}{HTML}{0065BD}
\definecolor{darkblue}{HTML}{001473}
\definecolor{checkmarkyes}{HTML}{1B9E77}
\definecolor{checkmarkno}{HTML}{D51965}
\usepackage[linktocpage,colorlinks=true,bookmarksnumbered=true,breaklinks=true,bookmarks=false,linkcolor=tumblue,citecolor=darkblue,urlcolor=darkblue]{hyperref}

\newfloatcommand{capbtabbox}{table}[][\FBwidth]
\DeclareCaptionLabelFormat{andfigure}{\figurename~\thefigure~\&~#1~#2}

\makeatletter
\DeclareRobustCommand\onedot{\futurelet\@let@token\@onedot}
\def\@onedot{\ifx\@let@token.\else.\null\fi\xspace}
\def\eg{\emph{e.g}\onedot} 
\def\ie{\emph{i.e}\onedot}

\def\wrt{w.r.t\onedot} 

\makeatother

\renewcommand{\v}[1]{\ensuremath{\mathbf{#1}}}  
\newcommand{\m}[1]{\ensuremath{\bm{#1}}}  

\newcommand{\Z}{\ensuremath{\mathbb{Z}}}

\newcommand{\R}{\ensuremath{\mathbb{R}}}

\newcommand{\round}[1]{\ensuremath{\left\lfloor{#1}\right\rceil}}
\newcommand{\wh}[1]{\widehat{#1}}

\newcommand{\p}[1]{\left(#1\right)}
\renewcommand{\b}[1]{\left\lbrace#1\right\rbrace}
\renewcommand{\u}{\textbf}

\newcommand{\pd}{\partial}  
\DeclareDocumentCommand{\pdd}{ O{} O{} m }{\frac{\pd^{#2}\!#1}{\pd#3^{#2}}}  

\DeclareMathOperator{\clip}{clip}
\DeclareMathOperator{\rank}{rank}

\newcommand{\ts}[1]{{\textsuperscript{#1}}}

\newcommand{\nf}[1]{{\small \textsf{#1}}}

\newcommand{\method}{{LR-QAT}}
\newcommand{\llama}{{LLaMA}}

\newcommand{\gc}[1]{\cellcolor{Gray}{#1}}
\newcommand{\gcb}[1]{\cellcolor{Gray}\textbf{#1}}

\newcommand{\ms}[2]{{#1}$^{\scriptsize \pm{}\text{#2}}$}

\newcommand{\da}{\ensuremath{\downarrow}}
\newcommand{\ua}{\ensuremath{\uparrow}}

\newcommand*{\boldcheckmark}{%
  \textpdfrender{
    TextRenderingMode=FillStroke,
    LineWidth=.5pt, 
  }{\checkmark}%
}
\newcommand{\boldyes}{\textcolor{checkmarkyes}{\boldcheckmark}}
\newcommand{\yes}{\textcolor{checkmarkyes}{\checkmark}}
\newcommand{\no}{\textcolor{checkmarkno}{\ding{53}}}
\newcommand{\oom}{{\ding{53}}}
\newcommand{\dcfn}{\ensuremath{\varphi}}  
\newcommand{\A}{\ensuremath{\m{A}}}
\newcommand{\B}{\ensuremath{\m{B}}}

\newcommand{\PhiO}{\ensuremath{\m{\Phi_0}}}
\newcommand{\s}{\ensuremath{\v{s}}}
\newcommand{\z}{\ensuremath{\v{z}}}
\newcommand{\sO}{\ensuremath{\v{s_0}}}
\newcommand{\zO}{\ensuremath{\v{z_0}}}
\newcommand{\W}{\ensuremath{\m{W}}}
\newcommand{\WO}{\ensuremath{\m{W_0}}}
\newcommand{\WZ}{\ensuremath{\m{W}_{\Z}}}
\newcommand{\What}{\ensuremath{\widehat{\m{W}}}}
\newcommand{\x}{\ensuremath{\v{x}}}

\newcommand{\xZ}{\ensuremath{\v{x}_{\Z}}}

\newcommand{\y}{\ensuremath{\v{y}}}

\title{Low-Rank Quantization-Aware Training for LLMs}

\author{Yelysei Bondarenko, Riccardo Del Chiaro, Markus Nagel \\
Qualcomm AI Research\thanks{\scriptsize Qualcomm AI Research is an initiative of Qualcomm Technologies, Inc.} \\
Amsterdam, The Netherlands\\
{\tt\small \{ybond, rdelchia, markusn\}@qti.qualcomm.com}\vspace{-.1cm}
}

\begin{document}

    \newif\ifappendix
    \newif\ifcontent
    
    \contenttrue     
    \appendixtrue    


\ifcontent
    \maketitle

    \begin{abstract}


Large language models (LLMs) are omnipresent, however their practical deployment is challenging due to their ever increasing computational and memory demands.
%
Quantization is one of the most effective ways to make them more compute and memory efficient.
%
Quantization-aware training (QAT) methods, generally produce the best quantized performance, however it comes at the cost of potentially long training time and excessive memory usage, making it impractical when applying for LLMs.
%
Inspired by parameter-efficient fine-tuning (PEFT) and low-rank adaptation (LoRA) literature, we propose~\u{\method} -- a lightweight and memory-efficient QAT algorithm for LLMs.
{\method} employs several components to save memory without sacrificing predictive performance: (a) low-rank auxiliary weights that are aware of the quantization grid; (b) a downcasting operator using fixed-point or double-packed integers and (c) checkpointing.
%
Unlike most related work, our method (i) is inference-efficient, leading to no additional overhead compared to traditional PTQ; (ii) can be seen as a general extended pretraining framework, meaning that the resulting model can still be utilized for any downstream task afterwards; (iii) can be applied across a wide range of quantization settings, such as different choices quantization granularity, activation quantization, and seamlessly combined with many PTQ techniques.
%
We apply {\method} to {\llama}-1/2/3 and Mistral model families and validate its effectiveness on several downstream tasks.
Our method outperforms common post-training quantization (PTQ) approaches and reaches the same model performance as full-model QAT at the fraction of its memory usage.
%
%
Specifically, we can train a 7B LLM on a single consumer grade GPU with 24GB of memory.
Our source code is available at~\url{https://github.com/qualcomm-ai-research/LR-QAT}.

\end{abstract}

    \section{Introduction}
\label{sec:intro}


In recent years, large language models (LLMs) have emerged as a powerful tool for a plethora of natural language processing tasks.
As these models continue to grow in size and capability, addressing their ever increasing computational and memory demands becomes crucial for practical deployment, especially when considering resource-constrained edge devices.

One of the most effective methods to tackle this problem is neural network quantization, which uses low-bit precision for weight and activation tensors.
%
While recent post-training quantization (PTQ) methods can help with decreasing the model size and improving the computational efficiency of LLMs, they typically lead to subpar performance, especially in the case of low-bit ($\leq 4$) quantization.
%
Quantization-aware training (QAT), conversely, yields significantly better model performance compared to PTQ. 
However, due to extreme model sizes of modern LLMs, using traditional QAT is very computationally expensive and requires a prohibitively high GPU memory usage, making it impractical.
%
%


Inspired by parameter-efficient fine-tuning (PEFT) and low-rank adaptation (LoRA) literature, we propose~\u{Low-Rank Quantization-Aware Training} (\u{\method}) -- a lightweight memory-efficient and inference-efficient QAT algorithm for LLMs.
%
{\method} reduces the memory requirements of training a 7B LLM from >98GB of GPU memory to <21GB without degrading the predictive performance compared to traditional full-model QAT, making it possible to train such models on a single consumer grade GPU.
Unlike most related work that combines low-rank adaptation with quantization, our method is also inference-efficient.
After the training is complete, the auxiliary matrices are naturally absorbed into the quantized weight tensor 
without loss of accuracy and no extra overhead at inference time. 
{\method} is positioned as a general~\emph{extended pretraining} method, as opposed to being strictly a fine-tuning method -- the resulting model is a low-bit general pretrained LLM, that can still be utilized for any task afterwards.
%
If needed, our resulting low-bit pretrained LLM can be fine-tuned on specific downstream tasks or used with multiple LoRA adapters for rapid switching between various tasks.

{\method} introduces and combines several innovations designed to reduce memory use without sacrificing model performance:
%
%
(1) a form of~\u{QAT with low-rank reparameterization}, in which we place the low-rank weights in the integer domain to ensure they align with the quantization grid of the pretrained weights. This allows for seamless fusion during inference into a single low-bit integer matrix.
%
%
%
(2) A~\u{downcasting operator} that represents the frozen pretrained weights as low-bit {\text{\nf{INT-}$b$}} ($b \leq 4$) double-packed into {\nf{INT8}} or as fixed-point values stored in {\nf{INT8}}.
%
(3) Finally, we combine the proposed quantization formulation with~\u{gradient checkpointing}  
%
to avoid aggressive memory spikes from storing some of the intermediate results in memory for the backward pass.

We apply {\method} to LLaMA-1/2/3 and Mistral model families and demonstrate its effectiveness on several general language modeling datasets and zero-shot evaluation on some of the common reasoning downstream tasks.
Our method outperforms recent PTQ approaches and reaches the same predictive performance as full-model QAT at the fraction of its memory usage.
%
Finally, our method can be applied across a wide range of quantization settings, including per-channel or per-block weight quantization, activation quantization, and can be combined with most of other PTQ techniques.
%


\section{Background and related work}
\label{sec:background}

\begin{figure}[tp]
    \centering

        \centering
        \includegraphics[width=0.36\textwidth]{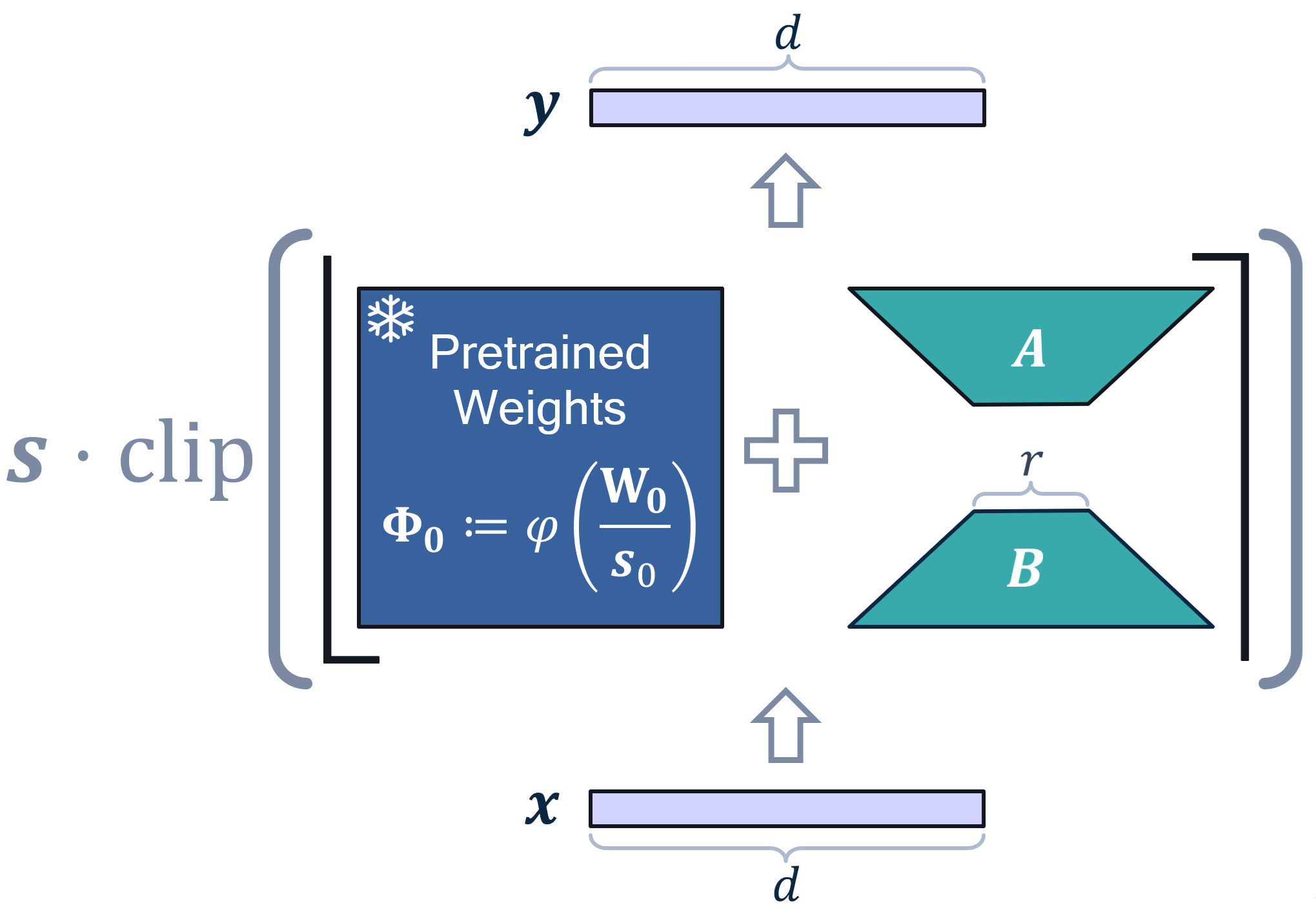}
        \hspace{0.03\textwidth}
        \centering
        \raisebox{-0.04\height}{\includegraphics[width=0.5725\textwidth]{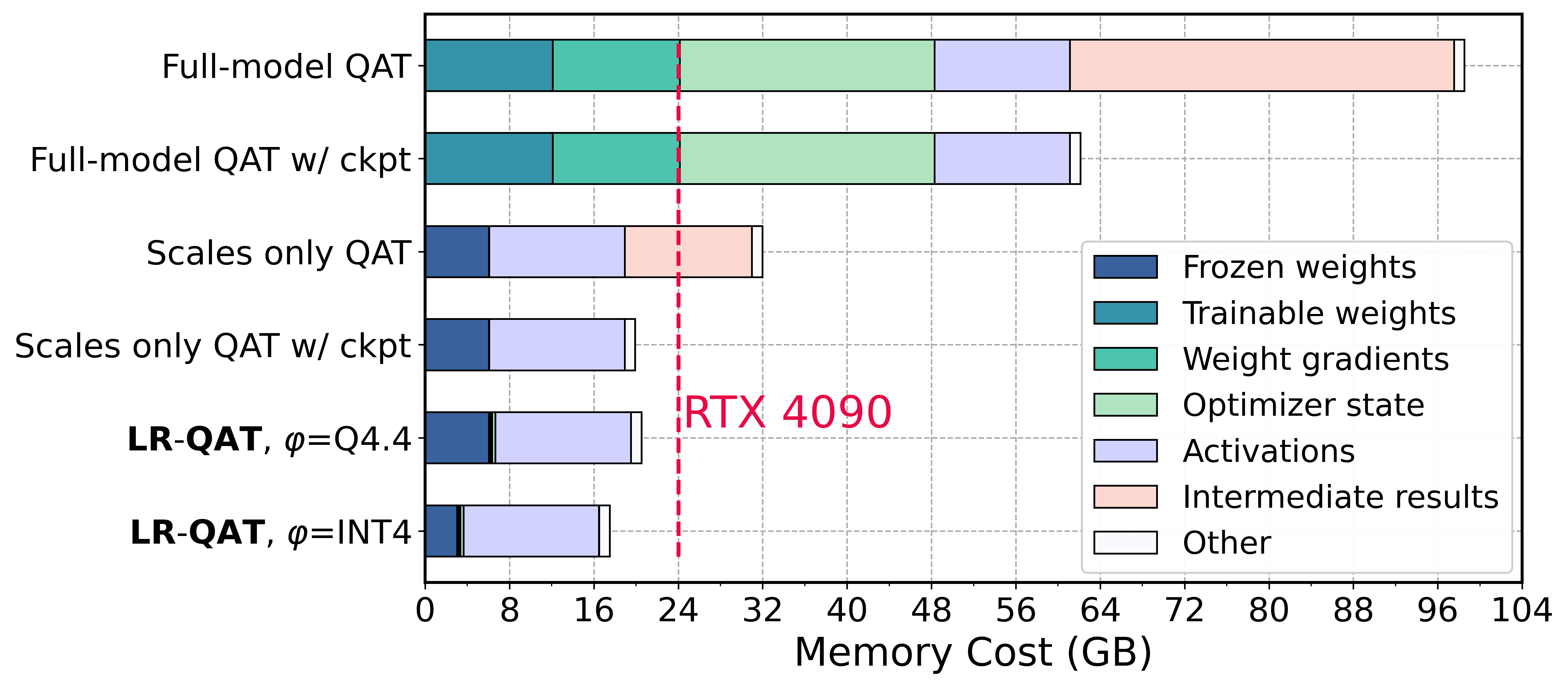}}
        \hspace{0.01\textwidth}  
        
        
    \caption{\emph{Left:} A schematic illustration of our proposed {\method}. $\x$ and $\y$ denote the input and the output of the linear layer.
    \emph{Right:} Memory requirements for training with various QAT techniques on {\llama}-2 7B, assuming batch size 1, sequence length 1024, $r=32$, and {\nf{BF16}} compute data type.
    `Intermediate results' refer to the results of some intermediate computations, \eg{} after rounding/clipping in~\eqref{eq:vanilla_qat}, which are saved in memory for the backward pass.
    }     
    \label{fig:new_fig_one}
    \vspace{-.2cm}
\end{figure}


Neural network quantization is one of the most powerful ways to reduce model footprint, data transfer and compute requirements.
By quantizing a model, high bit-width floating point weights and activations can be represented using low-bit numbers.
On top of that, by using low-bit fixed-point representations, such as~\nf{INT8}, one can further reduce energy consumption since the fixed-point operations are more efficient than their floating-point counterparts.
%
Quantizing to 8 bits or lower, however, typically introduces quantization noise in the model, resulting in a potential drop in accuracy/perplexity.

In this section, we provide a brief overview of uniform affine quantization and a summary of recent methods for LLM quantization.
We will discuss some of the trade-offs of those techniques.
Finally, we touch upon the challenges of LLM quantization and some of the limitations of current approaches.


\paragraph{Uniform affine quantization}
%
%
We use the following definition of the quantization function:
\begin{equation}
    \wh{\v{x}} := q\p{\v{x};\,s,z,b} = s \cdot 
    \vphantom{\Bigg(} \Big(\,\smash{\underbrace{\clip\!\p{\round{\frac{\x}{s}}+z;-2^{b-1}, 2^{b-1}-1}}_{\text{\normalsize $=: \xZ$}}} - z \Big),
    \label{eq:dequant}
\end{equation}

where $\v{x}$ denotes the quantizer input (i.e., network weights or activations), 
$s$ the higher precision quantization scale,
$z$ the integer zero offset, and $b$ the bitwidth.
$\round{\cdot}$ denotes the round-to-nearest-integer operator.
%
Quantization parameters $s$, $z$ can be shared across the components of $\x$.
%
One can see that such a quantizer approximates an original floating point vector as $\x \approx s \cdot \p{\xZ - z} $, where each element in $\xZ$ is a $b$-bit integer value.
This quantization scheme is called~\emph{uniform affine} or~\emph{asymmetric} quantization~\citep{hubara2017quantized,krishnamoorthi2018quantizing,zhou2016dorefa} and is one of the most commonly used quantization schemes because it allows for efficient implementation of fixed-point arithmetic.
In the case of~\emph{symmetric} quantization, we restrict the quantization grid to be symmetric around $z=0$.


\paragraph{Post-training quantization methods}
%
Post-training quantization (PTQ) algorithms take a pretrained high precision (\nf{FP32} / \nf{FP16} / \nf{BF16}) network and convert it directly into a fixed-point network without the need for the original training pipeline~\citep{banner2018post,cai2020zeroq,choukroun2019low,hubara2020improving,krishnamoorthi2018quantizing,li2021brecq,meller2019same,nagel2019data,adaround,zhao2019improving}.
%
%
These methods are either data-free or only require a small calibration dataset and are generally quite easy to use.
Having almost no hyperparameter tuning makes them usable via a single API call as a black-box method to quantize a pretrained neural network in a computationally efficient manner.

%
Post-training quantization of LLMs is a challenging task due to presence of numerical outliers in weights and activations~\citep{bondarenko-etal-2021-understanding,bondarenko2024quantizable,kovaleva2021bert,dettmersgpt3,sun2024massive}.
Existing LLM PTQ methods can be broadly categorized into~\emph{weights-only} quantization and~\emph{weight-activation} quantization.

Weights-only quantization focuses on converting weights to low-bit values.
%
For instance, GPTQ~\citep{frantar2022gptq} employs second-order information to iteratively round grouped weights and correct the quantization error in the remaining groups.
SpQR~\citep{dettmers2023spqr}, AWQ~\citep{lin2023awq} and OWQ~\citep{lee2024owq} emphasize the importance of so-called ``salient'' weights that correspond to high-magnitude activations.
Other recent W-only methods include~\citep{jeon2023frustratingly,lee2023flexround,luo2023long,chee2024quip}.

Weight-activation quantization compresses both weights and activations.
SmoothQuant~\citep{xiao2023smoothquant}, \texttt{LLM.int8()}~\citep{dettmersgpt3} and Outlier Suppression~\citep{wei2022outlier} achieve W8A8 quantization by managing activation outliers. \texttt{LLM.int8()} uses mixed-precision decomposition, while the other two employ channel-wise scaling.
OmniQuant~\citep{shao2023omniquant} modulates the extreme values of weights by optimizing the clipping threshold and shifts the challenge of quantization from activations to weights by employing the learnable equivalent transformation.
Some of the other recent W\&A PTQ methods are~\citep{lee2023enhancing,liu2023qllm,wei2023outlier,yuan2023rptq,tang2024easyquant,yao2022zeroquant,lin2024qserve}.


\paragraph{Quantization-aware training methods}
Quantization-aware training (QAT) methods~\citep{lsq+,lsq,gupta2015deep,jacob2018quantization,krishnamoorthi2018quantizing} simulate quantization during training, allowing the model to find more optimal solutions compared to PTQ approaches. 
However, better accuracy/perplexity comes at the cost of neural network training, i.e., longer training times, increased memory usage, need for labeled data and hyperparameter search.

The excessive training cost and memory usage of traditional QAT methods make them unsuitable for quantizing modern LLMs.
A few works that apply QAT to LLMs include LLM-QAT~\citep{liu2023llm} that combine QAT with data-free knowledge distillation, and EdgeQAT~\citep{shen2024edgeqat} that only considers tiny (sub 100M parameter) language models.


%
\paragraph{Low-rank adapters for fine-tuning}
%
Low-rank adaptation (LoRA)~\citep{hu2021lora} is a parameter efficient fine-tuning (PEFT) method that reduces memory requirements compared to standard training.
LoRA freezes the pretrained weights $\W = \WO$, and only trains a small set of low-rank trainable parameters, often termed~\emph{adapters}.
Given a linear projection $\y = \WO \x$ with $\WO \in \R^{m \times k}$, LoRA computes
\begin{flalign}
    \y = \WO \x + \frac{\alpha}{r} \A \B \x,
    \label{eq:lora}
\end{flalign}
where $\A\in\R^{m \times r}$, $\B\in\R^{r \times k}$, $r<\min\b{m, k}$ -- rank, and $\alpha$ is a scalar that is constant in $r$.
%
%
%
The benefits of LoRA are that it is much cheaper and often performs on par with or better than full fine-tuning. Additionally, the fine-tuned (floating-point) model can be deployed without extra cost, as the low-rank matrices can be fused into the pretrained weights after fine-tuning ($\W := \WO + \frac{\alpha}{r}\A\B$).

Naturally, there have been several works that explored the combination of LoRA and quantization.
%
QLoRA~\citep{dettmers2024qlora} quantizes the pretrained weights to 4 bit using (a non-uniform) {\nf{NF4}} format and dequantizes them in the forward pass to further reduce fine-tuning memory footprint.
%
%
QA-LoRA~\citep{xu2023qa} uses {\nf{INT4}} quantization and introduces
group-wise operators to enable quantization during inference
stage.
%
LoftQ~\citep{li2023loftq} proposed an iterative SVD-based procedure for initializing $\A$, $\B$ that yields faster fine-tuning convergence when used together with low-bit quantization.
LQ-LoRA~\citep{guo2023lq} further extends initialization technique from LoftQ to mixed precision and data aware cases.
Other recent works include~\citep{jeon2024l4q,zhang2024lqer}.

Finally, the closest work to ours is PEQA~\citep{kim2024memory}, that attempts to combine the benefits of inference-efficiency of QAT together with memory-efficiency of PEFT methods.
However, their approach is different since they focus on a task-specific fine-tuning as opposed to being a general extended pretraining method.
In addition to that, PEQA has significantly less degrees of freedom compared to our method, leading to a subpar performance.


%
\paragraph{Motivation}
%
While generally fast and simple, PTQ suffers from limited performance in low-bit scenarios.
%
%
%
Although QAT methods still perform well in low-bit regimes, their high training costs and memory usage make them impractical for LLMs.

LoRA-based methods address memory issues for efficient fine-tuning.
However, in most cases they don't consider efficient inference.
The adapters $\A$ and $\B$ are typically stored in higher precision formats, such as~\nf{BF16}. During inference, the frozen low-bit pretrained weights $\WO$ are dequantized to match the same data format, resulting in runtime overhead.

Simply quantizing adapters after training will lead to a different quantization grid compared to $\WO$, and quantizing them using the same quantization grid as $\WO$ will lead to a high error.
%
QA-LoRA is the only work we are aware of that attempts to fuse auxiliary LoRA weights back into the frozen $\WZ$.
%
Yet, their method is designed to only work with group-wise quantization with high number of groups (a small group size of $32$).
%
%
In addition to that, QA-LoRA and most of LoRA-based methods combine their proposed techniques with the task-specific fine-tuning, whereas we propose {\method} as an~\emph{extended pretraining} method.

We are inspired by LoRA-based methods to make QAT more memory- and runtime-efficient.
%
In addition to that, our goal is to design a method that is \emph{inference-efficient}, \ie~where the low-rank adapters can be fused back into a low-bit integer matrix $\WZ$ without any loss of accuracy/perplexity, yielding PTQ level of inference efficiency.
Contrary to QA-LoRA~\citep{xu2023qa}, we are not relaxing the quantization constraints -- our method is applicable at any weight quantization granularity.
%
Finally, we see our method as a general extended pretraining framework. 
The resulting model can afterwards still be used on any task.
We summarize different trade-offs for the discussed techniques in Table~\ref{tbl:02_methods_comparison}.
\begin{table}[tb]
    \setlength{\tabcolsep}{5pt}
    \centering
    \resizebox{0.72\columnwidth}{!}{%
        \begin{tabular}{ lcccc }
            \toprule
             Method & Accuracy & Memory efficiency & Inference efficiency \\
             \midrule       
             PTQ              &  \no & \yes & \yes \\
             (Full-model) QAT &  \yes & \no & \yes \\
             LoRA / PEFT  &  \yes & \yes & \no \\
             \midrule       
             \u{{\method} (ours)} &  \boldyes & \boldyes & \boldyes \\
             \bottomrule
        \end{tabular}
    } 
    \caption{A comparison between existing approaches and the proposed method.}
    \label{tbl:02_methods_comparison}
    \vspace{-.2cm}
\end{table}




    \section{Method}
\label{sec:method}


%
%
We now discuss the components of {\method} followed by a formal definition of {\method}.
%
%
%
\paragraph{QAT with low-rank adapters}
Let's recall how traditional QAT~\citep{lsq} works.
Given a linear layer with a weight matrix $\W \in \R^{m \times k}$ and assuming $b$-bit symmetric uniform affine quantization, the quantization is simulated as follows:
\begin{flalign}
    \What := \s \cdot \clip\p{\round{\frac{\W}{\s}}, -2^{b-1}, 2^{b-1}-1 },
    \label{eq:vanilla_qat}
\end{flalign}
%
where weights $\W$ are trainable parameters and the quantization scale $\s$ can be either fixed or also learned.
%
To be able to backpropagate through round-to-nearest operation in~\eqref{eq:vanilla_qat}, it is common to use~\emph{straight-through estimator} (STE, \citealt{bengio2013estimating}), where it is assumed that $\pdd[\round{t}]{t} = 1$.
%
%
When applied to LLMs, it is straightforward to see that this procedure is very expensive: we have to learn a comparable number of parameters that was used for pretraining, leading to excessive memory usage.

To make this approach more practical we~\emph{freeze} the pretrained weights $\W$ (denote $\WO$) and introduce low-rank adapters $\A\in\R^{m \times r}$, $\B\in\R^{r \times k}$, $r\ll\min\b{m, k}$.
%
We have to be careful where exactly those adapters are placed.
As discussed in Section~\ref{sec:background}, after the training is complete, we want $\A$ and $\B$ to be seamlessly integrated into a single $b$-bit integer matrix $\WZ$ without loss of accuracy to facilitate efficient inference.
%
To accommodate that, we put the auxiliary matrices inside the rounding operator as follows
\begin{flalign}
    \What := \s \cdot \clip\p{\round{\frac{\WO}{\s} + \frac{\alpha}{r}\A\B}, -2^{b-1}, 2^{b-1}-1 },
    \label{eq:low_rank_qat}
\end{flalign}
where we are using STE assumption for the rounding operation to compute the gradients of the loss {\wrt} $\A$, $\B$ and $\s$. We further employ a scaling factor $\alpha / r$ used in LoRA~\citep{hu2021lora} to reduce the need to retune hyperparameters as we vary the rank $r$.
After training is complete,~\eqref{eq:low_rank_qat} can be represented as regular fixed point tensor, $\What = \s \cdot \WZ$, without any extra effort or loss of accuracy and therefore enabling efficient inference without any extra overhead.
%
Note that this is different to most of the literature, such as QLoRA~\citep{dettmers2024qlora}, where adapters are placed outside of the quantization function (such as $\y = \What \x + \A\B\x$) and are typically stored in higher precision formats such as~\nf{BF16}.


\paragraph{Downcasting operator}
%
The formulation~\eqref{eq:low_rank_qat} is already significantly more memory efficient compared to standard full-model QAT~\eqref{eq:vanilla_qat}. 
%
We don't need to compute neither gradients {\wrt} weights $\W$ nor the respective first or second-order momentum terms for Adam-based optimizers, and only need to do so for the auxiliary matrices $\A$ and $\B$, which is noticeably more affordable provided $r \ll \min\!\b{m, k}$.

Given that the weight matrix $\WO$ is frozen, the next natural step to further reduce the memory utilization is to store it in a lower-precision format.
%
%
One could directly apply downcasting to $\WO$ in~\eqref{eq:low_rank_qat}. 
However, it's important to note that these weights are divided by the scale $\s$ during every forward pass. 
To ensure stable training, the scale generally needs to be stored in a high-precision format.
%
Therefore, to simplify further, we propose the following variant of low-rank QAT:
\begin{flalign}
    \What := \s \cdot \clip\p{\round{\frac{\WO}{\sO} + \frac{\alpha}{r}\A\B}, -2^{b-1}, 2^{b-1}-1 },
    \label{eq:low_rank_qat_with_s0}
\end{flalign}
where we use the initial scale\footnote{A frozen scale obtained after initial range estimation before the training begins.}
$\sO$ instead of learned scale $\s$ inside the rounding operator, and the rest is the same as in~\eqref{eq:low_rank_qat}.
Now the entire fraction $\WO / \sO$ is fixed and we can store it in a lower-precision format.
Note that the scale $\s$ outside of the clipping operator can still be learned.
%
Empirically, we found that~\eqref{eq:low_rank_qat_with_s0} performs consistently on par with or even slightly better compared to~\eqref{eq:low_rank_qat}.

During training the pretrained weights are represented and stored as follows
\begin{flalign}
    \PhiO := \dcfn\!\p{\frac{\WO}{\sO}},
    \label{eq:phi}
\end{flalign}
where $\dcfn\!\p{\cdot}$ is a~\emph{downcasting operator} that encapsulates a choice of different numeric formats or other preprocessing computations.
%
In the simplest form, $\dcfn\!\p{\cdot}$ would cast the input to one of pre-existing floating-point formats, such as~\nf{FP16}, \nf{BF16}, \nf{FP8} etc.

Inspired by traditional fixed point quantization, we also explore integer representations for $\dcfn\!\p{\cdot}$.
Specifically, $\dcfn\!\p{x} = \clip\p{ \round{x}, -2^{b-1}, 2^{b-1}-1}$ corresponds to a standard $b$-bit integer quantization and can be stored as {\nf{INT-}$b$} number.
We denote this approach $\dcfn = {\text{\nf{INT-}$b$}}$ for brevity.
%
In addition to that, in case of low-bit quantization ($b \leq 4$), which is a primary focus of this work, 
two {\nf{INT-}$b$} numbers can be~\emph{double-packed}
into a single~\nf{INT8} number, leading to further memory savings.
This is helpful because most of the common deep learning frameworks like PyTorch, at the time of writing this paper, don't natively support low-bit formats such as~\nf{INT4} yet.

Using $\dcfn = {\text{\nf{INT-}$b$}}$ naturally leads to aggressive memory reduction by only keeping the integer part of (clipped) $\WO / \sO$.
In our preliminary experiments, we found that this setting, combined with the standard initialization for $\A$ and $\B$ used in~\citep{hu2021lora}, did not work as well compared to $\dcfn = {\text{\nf{BF16}}}$.
%
This indicates the importance of keeping some information of the fractional part of $\WO / \sO$ and potentially the need for better initialization of auxiliary matrices.

We address this problem in two distinct ways:
We adapt and experiment with a variant of SVD-based initialization for low-rank matrices $\A$, $\B$ proposed in~\citep{li2023loftq} before we apply a downcasting operator to $\WO / \sO$, to capture some of the information of it's fractional part.
With this approach we can still employ a double-packing since we are still using $\dcfn = {\text{\nf{INT-}$b$}}$.

Another way is to use {\nf{INT8}} storage type, allocate $b$ bits to represent the integer part as before, but utilize the remaining $8-b$ bits for storing the approximate fractional part ($2\leq b \leq 7$).
In other words, we represent $\PhiO$ using fixed-point numbers.  
%
Assuming the rest of the computation is performed in {\nf{BF16}}, we define the downcasting and the corresponding upcasting operators as follows:
%
\begin{equation}
\begin{aligned}
    \dcfn\!\p{x}      &= \text{\nf{INT8}}\!\p{\round{2^{8-b} \cdot \clip\p{x, -2^{b-1}, 2^{b-1}-1}}}, \label{eq:phi_fixed_point} \\
    \dcfn^{-1}\!\p{x} &= \text{\nf{BF16}}\!\p{x}\!/{2^{8-b}}.
\end{aligned}
\end{equation}
A fixed-point number where $n$ bits are used for the integer part of the value and $m$ bits are used for the fractional part are commonly denoted~\citep{oberstar2007fixed} as {\nf{Qn.m}}.
For brevity, we will refer to~\eqref{eq:phi_fixed_point} as $\dcfn = {\text{\nf{Q$b$.$(8-b)$}}}$.
%
In this work we will be mainly focusing on $b \in \b{3, 4}$, which corresponds to {\nf{Q3.5}} and {\nf{Q4.4}}, respectively.


%
\paragraph{Gradient checkpointing}
Note that both in the original LoRA paper~\citep{hu2021lora} and in the related work like QLoRA~\citep{dettmers2024qlora}, there is no need to compute the product $\A\B$ explicitly.
%
Instead, those matrices are multiplied with the activations $\x$ as $\A\p{\B\x}$.
However, we have to compute a product $\A\B$ in~\eqref{eq:low_rank_qat_with_s0}, and in the na\"ive implementation of our method, this product together with the results of some intermediate computations (e.g., after rounding and clipping) will be automatically kept in memory for the backward pass, leading to increased memory usage.
%
To prevent this, we employ gradient checkpointing~\citep{chen2016training} on~\eqref{eq:low_rank_qat_with_s0}.
In other words, we recompute the quantizer function in the backward pass, leading to a slight runtime overhead but avoiding significantly increased memory usage.


\paragraph{\method}
Using the components described above, we define {\method} for a single layer with a (pretrained) weight matrix $\WO$ as follows
\begin{flalign}
    \What := \s \cdot \clip\p{\round{\PhiO + \frac{\alpha}{r}\A\B}, -2^{b-1}, 2^{b-1}-1 },
    \label{eq:lr_qat}
\end{flalign}
where $\s$ -- trainable or frozen quantization scale with the initial value of $\sO$, $\A$, $\B$ -- trainable rank $r$ auxiliary matrices, $\PhiO := \dcfn\!\p{\WO / \sO}$ -- frozen representation of the original pretrained weights, and $\varphi$ is the downcasting operator.
To avoid excessive memory allocation for the results of intermediate computations in~\eqref{eq:lr_qat} involving the product $\A\B$, we apply checkpointing on $\What$.
After the training is complete, low-rank adapters are naturally integrated into a single integer matrix $\WZ = \clip\p{\cdots}$ without loss of accuracy.
Note, while we presented our method for symmetric quantization which is commonly used for weights~\citep{quantization_whitepaper}, it can equally be applied for asymmetric quantization by adding a zero offset $z$ outside the rounding operation as shown in~\eqref{eq:dequant}.





%
%



    \section{Experiments}
\label{sec:experiments}
 
%
%
We assess the effectiveness of {\method} by conducting experiments on {\llama} 7B~\citep{touvron2023llama}, {\llama}-2 7B/13B~\citep{touvron2023llama2}, {\llama}-3 8B~\citep{llama3}, and Mistral-0.1 7B~\citep{jiang2023mistral}.
%
%
We first explore the impact of the choice of rank $r$, a downcasting operator $\dcfn\!\p{\cdot}$, and the initialization of auxiliary matrices $\A$, $\B$.
We then compare our method in terms of accuracy to standard full-model QAT, other baselines, and the related work.
All detailed hyperparameters of our experiments are in Appendix~\ref{app:exp_details}.

\paragraph{Quantization}
We experiment with both weight-only and weight-activation quantization. 
The default settings are {\nf{INT4}}\,/\,{\nf{INT3}}\,/\,{\nf{INT2}} per-channel (denoted `pc') and group-wise weight quantization with a group size of 128 (denoted `g128').
We use symmetric quantization, except the {\nf{INT2}} case, where we use asymmetric quantization~\eqref{eq:dequant}, for a fair comparison with related work.
We quantize all linear layers, except the classification head.
In weight-activation quantization, defaults are {\nf{INT4}} per-channel weight and
per-token activation quantization~\citep{dettmersgpt3}.
Following OmniQuant~\citep{shao2023omniquant}, we quantize all inputs to matmuls with exception of the softmax output and additionally quantize the KV-cache as in LLM-QAT~\citep{liu2023llm}.

\paragraph{Datasets and training}
%
%
We apply our method to all linear layers in the attention blocks, both in self-attention and in the feed-forward network.
We only train low-rank auxiliary matrices $\A$, $\B$ and the quantization parameters $\s$ and keep embedding layers, final classification head and RMSNorm parameters frozen.
In the case of asymmetric weight quantization, a zero offset $\z$ is set during range estimation phase and kept frozen throughout training ($\z=\zO$).

We train on a small subset of SlimPajama~\citep{slimpajama}, 
which is an open-source dataset similar to the original one used for pretraining {\llama} models.
In all experiments we train using batch size $32$ and a maximum sequence length of $1024$.
For all weight-only and weight-activation quantization results, we train for $10^4$ steps.
For ablation studies in Sections~\ref{subsec:ablation_rank} and~\ref{subsec:ablation_phi_init_dtype} we use shorter training of $10^3$ steps.
%
%
%
%
%
We select hyperparameters based on the perplexity of a small subset of Wikipedia validation set (512 sequences).

\paragraph{Evaluation}
Following the previous work~\citep{frantar2022gptq,xiao2023smoothquant,shao2023omniquant,liu2023llm}, we evaluate quantized models by reporting the perplexity of language generation on WikiText-2~\citep{merity2016pointer}, using a sequence length of $2048$.
%
%
We also report zero-shot accuracy on a set of common sense reasoning tasks including BoolQ~\citep{clark2019boolq}, PIQA~\citep{bisk2020piqa}, Winogrande~\citep{sakaguchi2021winogrande}, ARC~\citep{clark2018think}, and HellaSwag~\citep{zellers2019hellaswag}.
For zero-shot evaluation, we use the LM Evaluation Harness framework~\citep{gao2021framework}.
Specifically, we use lm\_eval v0.4.2 and report~\texttt{acc\_norm} for tasks where it's available (PIQA, ARC-e, ARC-c, HellaSwag) and otherwise~\texttt{acc} (BoolQ and Winogrande).

\paragraph{Baselines}
%
We compare with round-to-nearest quantization (RTN), where we set the ranges based on minimizing the $L^p$-norms between quantized and unquantized weights and report the best performing configuration.
We also use that as initialization for {\method}.
%
For weight-only quantization, we compare with GPTQ~\citep{frantar2022gptq}, AWQ~\citep{lin2023awq}, and OmniQuant~\citep{shao2023omniquant}.
We also compare with our implementation of PEQA~\citep{kim2024memory} and full-model QAT (LSQ)~\citep{lsq}, where we follow the same experimental setup as for our method, together with the best RTN initialization, for a fair comparison.
%

%
%
%
For weight-activation quantization, we compare our method with RTN, SmoothQuant~\citep{xiao2023smoothquant}, LLM-QAT~\citep{liu2023llm}, Outlier Suppression$+$~\citep{wei2023outlier}, OmniQuant~\citep{shao2023omniquant}, and our implementation of PEQA~\citep{kim2024memory}.
Following~\citep{liu2023llm}, we compare to them in several different settings, where the weights, activations and KV cache values are quantized to different bitwidths (denoted as W-A-KV).
%


%
%

%
%
\subsection{The impact of rank $r$}
\label{subsec:ablation_rank}
\begin{table*}[t]
\begin{floatrow}

%
%
    
    \includegraphics[width=0.308\textwidth]{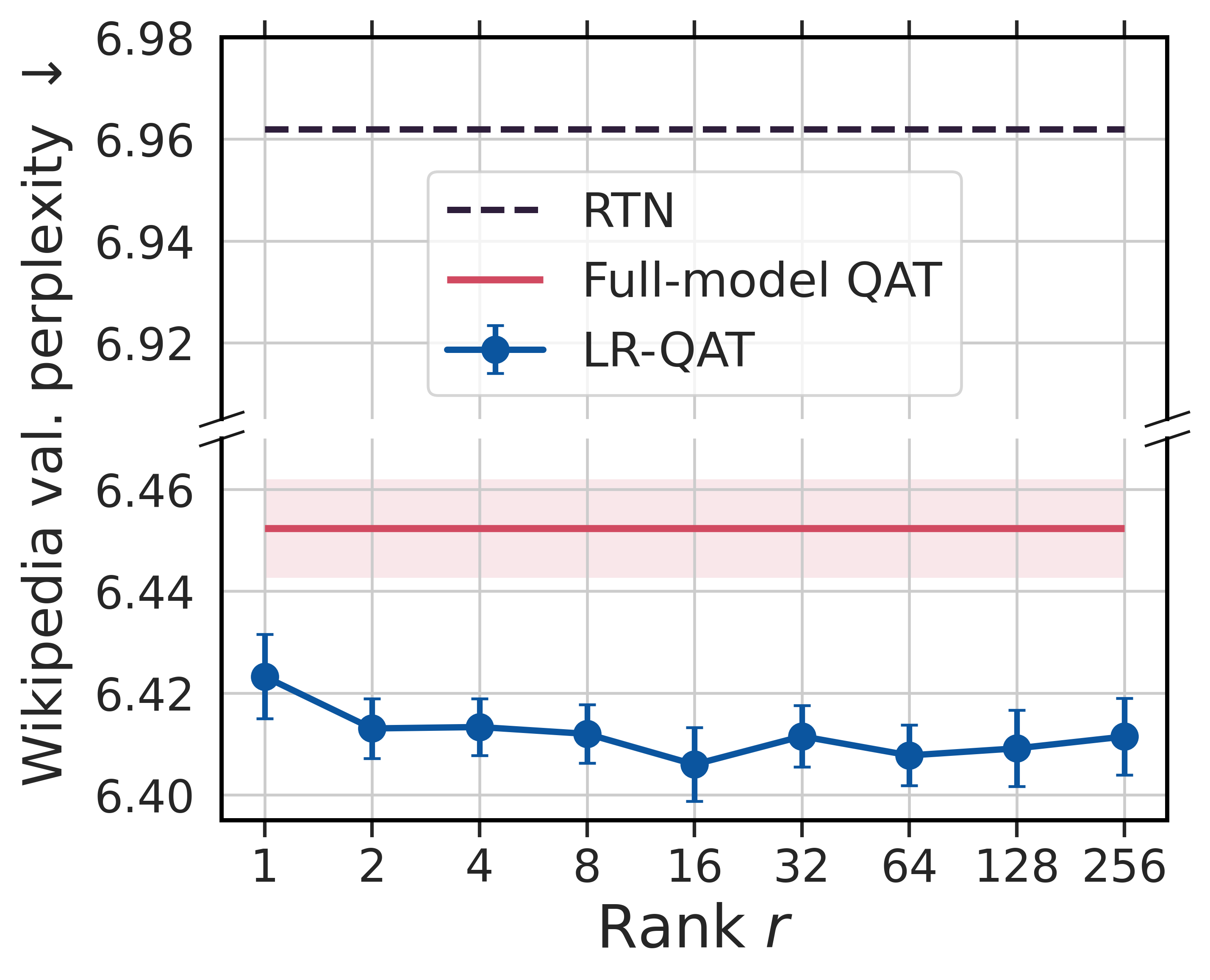}
    \hspace{0.03\textwidth}
    %

    \capbtabbox{%
        \resizebox{0.64\textwidth}{!}{%
        \begin{tabular}{ lllcccc }
            \toprule
            \multicolumn{1}{c}{\multirow{2}{*}{$\dcfn\!\p{\cdot}$}} & \multirow{2}{*}{dtype} & \multicolumn{1}{c}{\multirow{2}{*}{$\A$, $\B$ init.}} & \multicolumn{2}{c}{WikiText-2 $\da$} & \multicolumn{2}{c}{Zero-shot acc. $\ua$} \\[0.1em]
            & & & W4\,pc & W3\,pc & W4\,pc & W3\,pc \\
            \midrule
            {\nf{FP32}} & {\nf{FP32}} & LoRA & 5.69 & 6.21 & 69.28 & 66.62 \\
            \midrule
            {\nf{FP16}} & {\nf{FP32}} & LoRA & $+$0.00 & \u{$+$0.01} & $-$0.13 & $-$0.01 \\
            {\nf{BF16}} & {\nf{FP32}} & LoRA & \u{$-$0.01} & \u{$+$0.01} & $+$0.11 & \u{$+$0.45} \\
            {\nf{Q4.4} / \nf{Q3.5}} & {\nf{FP32}} & LoRA & \u{$-$0.01} & \u{$+$0.01} & \u{$+$0.16} & $+$0.31 \\
            \gcb{\nf{Q4.4} / \nf{Q3.5}} & \gcb{\nf{BF16}} & \gcb{LoRA} & \gcb{$-$0.01} & \gcb{$+$0.01} & \gc{$+$0.15} & \gc{$+$0.31} \\
            {\nf{INT-4} / \nf{INT-3}} & {\nf{FP32}} & LoRA & $+$0.02 & $+$20.5 & $-$0.04 & $-$22.8 \\
            {\nf{INT-4} / \nf{INT-3}} & {\nf{FP32}} & LoftQ ($T=1$) & $+$0.28 & $+$0.18 & $-$0.67 & $+$0.26 \\
            {\nf{INT-4} / \nf{INT-3}} & {\nf{FP32}} & LoftQ ($T=64$) & $+$0.40 & $+$1.37 & $-$1.40 & $-$2.01 \\
            \bottomrule
        \end{tabular}
        } 
    }  
    
    \captionlistentry[figure]{}
    \label{fig:04_ablation_rank}
    \vspace{-.1cm}
    \captionsetup{labelformat=andfigure}
    \caption{%
    \emph{Left}: The performance of {\method} ($\dcfn{} = {\text{\nf{Q4.4}}}$) depending on the rank $r$ of auxiliary matrices $\A$ and $\B$ on {\llama}-2 7B with W4 per-channel quantization.
    We report mean and standard deviation over 5 runs with different random seeds.
    \emph{Right}: The performance of {\method} applied to {\llama}-2 7B depending on the choice of downcasting operator $\dcfn\!\p{\cdot}$, compute data type, and initialization method for low-rank auxiliary matrices.
    We report WikiText-2 test set perplexity, lower is better, and average zero-shot accuracy of 6 tasks, higher is better.
    Numbers marked in bold are the best results.
    }
    \label{tbl:04_ablation_dtype_init}
\end{floatrow}
\vspace{-.1cm}
\end{table*}

We investigate the effect of different values of rank $r$ of the auxiliary matrices $\A$ and $\B$ and present results in Figure~\ref{fig:04_ablation_rank}.
%
%
%
Increasing the rank from $1$ to $32$ leads to progressively slightly better performance, excluding one outlier.
%
The fact that using $r>32$ doesn't lead to further improvement in perplexity is likely because of the limited number of training steps we used for this experiment ($10^3$), and that more steps might be needed for the procedure to fully converge.
%
Interestingly, a rank $r$ as small as 1 already performs really well.
We hypothesize that this is the case because of the following.
Even though $\rank\!\p{\A\B}=1$, by applying a low-rank approximation inside the rounding and clipping operators in~\eqref{eq:lr_qat}, this can overall lead to a high-rank perturbation to the original weights $\PhiO$ (in the integer domain).
Finally, for all ranks we observe only a small standard deviation between $0.005$ and $0.008$ ppl., indicating the robustness of {\method} to a random initialization of $\B$.
Going forward, we use $r=32$ in all our experiments\footnote{This amounts to only 1.2\% of the total number of parameters for 7B {\llama} model.}.

%
\subsection{The choice of the downcasting operator $\dcfn\!\p{\cdot}$ and $\A$, $\B$ initialization}
\label{subsec:ablation_phi_init_dtype}
%
%
We study the effect of several choices of the downcasting operators discussed in Section~\ref{sec:method} and summarize results in Table~\ref{tbl:04_ablation_dtype_init}.
%
We can see that by going from {\nf{FP32}} to {\nf{BF16}}, and finally to an 8-bit fixed-point representation of $\PhiO$, aside from memory savings we also maintain the same WikiText-2 perplexity and even slightly improve zero-shot accuracy.
%
The latter is likely due to a slight regularization effect caused by the fact that we discard some of the information in the fractional part in $\WO / \sO$, some of which might be noise.
%
One step further, however, while $\dcfn{} = {\text{\nf{INT-$b$}}}$ still leads to a good model performance in the case of 4-bit weight quantization, it completely breaks for W3.

So far, we initialized matrices $\A$ and $\B$ following the procedure proposed in LoRA~\citep{hu2021lora} where $\B$ is initialized to zero, and $\A$ is initialized randomly as in~\citep{he2015delving}.
%
We refer to this initialization scheme as `LoRA'.
We hypothesize that a poor performance of $\dcfn{} = {\text{\nf{INT3}}}$ can be explained by the fact that we lose all the information in the fractional part of $\WO / \sO$ and that without that information it is difficult for low-rank approximation to learn.
To address this, we adapt and experiment with a variant of SVD-based initialization proposed in LoftQ~\citep{li2023loftq}.
We see that using LoftQ initialization with $T=1$ step recovers almost all the predictive performance compared to a fixed-point representation.
Increasing number of LoftQ steps, or applying it to a 4-bit case, however, did not help.

Finally, when using the fixed point representation for $\PhiO$, we still maintain the same model performance by switching the compute data type\footnote{A data type used for activations, gradients, and frozen parameters.} from {\nf{FP32}} to {\nf{BF16}}, where the latter is what is commonly used for LLMs.
Going forward, we use $\dcfn = {\text{\nf{Q$b$.$(8-b)$}}}$ with `LoRA' initialization and $\nf{BF16}$ compute data type.

\subsection{Comparison with full-model QAT}
\label{subsec:comparison_with_full_model_qat}
\begin{table*}[t]
    \setlength{\tabcolsep}{6pt}
    \centering
    \caption{A comparison of the proposed method ($\dcfn{} = {\text{\nf{Q4.4}}}$) with the full-model QAT on {\llama}-2 7B with W4 and W3 per-channel quantization.
    We report mean and standard deviation over 5 runs with different random seeds.
    We also report the maximum GPU memory with (without) gradient checkpointing and the runtime on a Nvidia A100 80GB GPU.
    }
    \label{tbl:04_comparison_with_vanilla_qat}
    \resizebox{1.0\columnwidth}{!}{%
    \begin{tabular}{ lcccccc }
        \toprule
        \multicolumn{1}{c}{\multirow{2}{*}{Method}} & {GPU mem.,} & {Time/100 steps,} & \multicolumn{2}{c}{WikiText-2 ppl. $\da$} & \multicolumn{2}{c}{Zero-shot acc. $\ua$} \\[0.05em]
        & {\small GB} & {sec} & W4\,pc & W3\,pc & W4\,pc & W3\,pc \\
        \midrule
        Full-model QAT (LSQ) & {62.2 (98.5)} & {\ms{3248}{7}} & {\ms{5.77}{0.02}} & {\ms{6.14}{0.01}} & {\ms{68.96}{0.29}} & {\ms{67.14}{0.13}} \\
        \gcb{\method{} (ours)} & \gcb{20.5} & \gcb{\ms{1522}{5}} & \gcb{\ms{5.66}{0.00}} & \gcb{\ms{6.13}{0.02}} & \gcb{\ms{69.72}{0.32}} & \gcb{\ms{67.70}{0.25}} \\
        \bottomrule
    \end{tabular}
    }  
    \vspace{-.2cm}
\end{table*}
Finally, before presenting our main set of results, we compare our method with a standard full-model QAT (LSQ)~\citep{lsq}.
For full-model QAT, we follow the same training setup as for our method.
We also tune the maximum value of the learning rate for $\W$ using the following search space $\b{\text{1e-5, \u{5e-5}, 1e-4, 5e-4, 1e-3}}$ and select the best configuration based on Wikipedia validation perplexity.
Note that we use the same learning rate for $\s$ for both full-model QAT and our method.

As we can see in Table~\ref{tbl:04_comparison_with_vanilla_qat}, training with our method leads to on par or better predictive performance at a significantly lower memory usage and runtime compared to full-model QAT.
We include results for other models and bitwidths in Table~\ref{tbl:04_weight_only_quant_results}.
A more detailed runtime comparison can also be found in Table~\ref{tbl:A_runtime}.


\subsection{Main results}
\begin{table*}[t]
    \setlength{\tabcolsep}{6pt}
    \centering
    \caption{\textbf{Weight-only quantization results for {\llama} and Mistral models}. 
    We report WikiText-2 test set perplexity (lower is better) and average zero-shot accuracy (higher is better). Models marked `L1'/`L2'/`L3', and `M' denote {\llama}-1/2/3 and Mistral, respectively. Numbers marked in bold are SOTA or on par (within $0.05$). {\ts{\S}}Uses asymmetric weight quantization. {\oom} denotes out of memory.}
    \label{tbl:04_weight_only_quant_results}
    \renewcommand{\arraystretch}{1.0}
    \resizebox{0.975\columnwidth}{!}{%
        \begin{tabular}{clccccc!{\color{Gray2}\vline}ccccc}
            \toprule
            \multirow{2}{*}{\# Bits} & \multicolumn{1}{c}{\multirow{2}{*}{Method}} & \multicolumn{5}{c}{WikiText-2 perplexity $\da$} & \multicolumn{5}{c}{Avg. zero-shot accuracy $\ua$} \\[0.25em]
              & & L1-7B & L2-7B & L2-13B & L3-8B & M-7B & L1-7B & L2-7B & L2-13B & L3-8B & M-7B \\
            \midrule
            FP16 &  & 5.68 & 5.47 & 4.88 & 6.14 & 5.25 & 69.68 & 70.47 & 73.18 & 74.22 & 75.69 \\
            \midrule
            \multirow{6}{*}{\shortstack{W4\,pc}} 
            & RTN & 6.33 & 6.14 & 5.21 & 7.53 & 5.91 & \u{68.51} & 68.88 & 71.73 & 72.19 & 73.44 \\
            & GPTQ\ts{\S} & 6.13 & 5.83 & 5.13 & - & - & 64.95 & - & - & - & - \\
            & AWQ & 6.08 & 6.15 & 5.12 & - & - & - & - & - & - & - \\
            & OmniQuant\ts{\S} & \u{5.86} & 5.74 & \u{5.02} & - & - & - & - & - & - & - \\
            & LSQ (our impl.) & 5.94 & 5.77 & {\oom} & 6.87 & 5.73 & 68.37 & 68.96 & {\oom} & 73.28 & 72.88 \\
            & PEQA (our impl.) & \u{5.86} & \u{5.71} & \u{5.03} & 7.51 & 5.56 & \u{68.49} & 69.23 & 72.51 & 72.79 & 73.73 \\
            & \gcb{LR-QAT (ours)} & \gcb{5.84} & \gcb{5.66} & \gcb{5.03} & \gcb{6.78} & \gcb{5.46} & \gcb{68.54} & \gcb{69.72} & \gcb{73.19} & \gcb{73.84} & \gcb{74.44} \\
            \midrule
            \multirow{6}{*}{\shortstack{W4\,g128}} 
            & RTN & 6.05 & 5.78 & 5.04 & 6.96 & 5.49 & 68.93 & 69.75 & 72.94 & 72.30 & 75.07 \\
            & GPTQ\ts{\S} & 5.85 & \u{5.61} & \u{4.98} & - & - & - & - & - & - & - \\
            & AWQ & 5.81 & \u{5.62} & \u{4.97} & - & - & - & - & - & - & - \\
            & OmniQuant\ts{\S} & \u{5.77} & \u{5.58} & \u{4.95} & - & - & - & - & - & - & - \\
            & LSQ (our impl.) & \u{5.76} & \u{5.61} & {\oom} & \u{6.58} & 5.67 & \u{69.17} & 69.68 & {\oom} & 73.31 & 72.90 \\
            & PEQA (our impl.) & \u{5.75} & 5.67 & 5.02 & 6.89 & 5.48 & \u{69.19} & 69.64 & 72.80 & 72.99 & 73.34 \\
            & \gcb{LR-QAT (ours)} & \gcb{5.75} & \gcb{5.59} & \gcb{4.97} & \gcb{6.57} & \gcb{5.37} & \gcb{69.15} & \gcb{69.88} & \gcb{72.91} & \gcb{73.66} & \gcb{75.28} \\
            \midrule    
            \multirow{6}{*}{\shortstack{W3\,pc}} 
            & RTN & 12.88 & 26.73 & 8.71 & 34.10 & 9.49 & 54.66 & 43.87 & 55.01 & 47.46 & 64.58 \\
            & GPTQ\ts{\S} & 8.06 & 8.37 & 6.44 & - & - & - & - & - & - & - \\
            & AWQ & 11.88 & 24.00 & 10.45 & - & - & - & - & - & - & - \\
            & OmniQuant\ts{\S} & 6.49 & 6.58 & \u{5.58} & - & - & - & - & - & - & - \\
            & LSQ (our impl.) & \u{6.29} & \u{6.14} & {\oom} & \u{8.14} & \u{6.06} & 66.29 & 67.14 & {\oom} & 69.58 & 71.61 \\
            & PEQA (our impl.) & 6.56 & 6.45 & 5.73 & 26.20 & 6.51 & 65.75 & 65.44 & 69.81 & 51.05 & 71.02 \\
            & \gcb{LR-QAT (ours)} & \gcb{6.27} & \gcb{6.13} & \gcb{5.54} & \gcb{8.12} & \gcb{6.03} & \gcb{66.60} & \gcb{67.70} & \gcb{71.22} & \gcb{70.46} & \gcb{71.87} \\
            \midrule
            \multirow{6}{*}{\shortstack{W3\,g128}} 
            & RTN & 7.96 & 7.61 & 6.20 & 15.11 & 6.77 & 63.50 & 63.20 & 67.60 & 57.74 & 69.35 \\
            & GPTQ\ts{\S} & 6.55 & 6.29 & 5.42 & - & - & - & - & - & - & - \\
            & AWQ & 6.46 & 6.24 & \u{5.32} & - & - & - & - & - & - & - \\
            & OmniQuant\ts{\S} & \u{6.15} & \u{6.03} & \u{5.28} & - & - & - & - & - & - & - \\
            & LSQ (our impl.) & \u{6.20} & \u{6.02} & {\oom} & 8.08 & 5.90 & 66.53 & 68.36 & {\oom} & 70.11 & 71.96 \\
            & PEQA (our impl.) & 6.22 & 6.05 & 5.58 & 9.64 & \u{5.85} & 66.66 & {68.10} & 70.29 & 67.19 & 72.21 \\
            & \gcb{LR-QAT (ours)} & \gcb{6.17} & \gcb{5.98} & \gcb{5.32} & \gcb{7.74} & \gcb{5.80} & \gcb{66.81} & \gcb{68.62} & \gcb{71.51} & \gcb{70.48} & \gcb{72.41} \\
            \midrule
            \multirow{6}{*}{\shortstack{W2\,pc}} 
            & RTN\ts{\S} & 4.9e3 & 5.2e3 & 5.2e3 & 6.4e4 & 6.8e3 & 37.92 & 36.52 & 36.27 & 36.80 & 36.59 \\
            & GPTQ\ts{\S} & 2.1e3 & 7.7e3 & 2.1e3 & - & - & - & - & - & - & - \\
            & OmniQuant\ts{\S} & 15.47 & 37.37 & 17.21 & - & - & - & - & - & - & - \\
            & PEQA (our impl.)\ts{\S} & 8.24 & 9.34 & 7.51 & 14.47 & 8.08 & 59.83 & 57.40 & 62.29 & 57.72 & 64.48 \\
            & \gcb{LR-QAT (ours)\ts{\S}} & \gcb{7.99} & \gcb{8.51} & \gcb{7.16} & \gcb{12.52} & \gcb{7.98} & \gcb{61.77} & \gcb{60.03} & \gcb{65.28} & \gcb{58.49} & \gcb{65.11} \\
            \midrule
            \multirow{7}{*}{\shortstack{W2\,g128}} 
            & RTN\ts{\S} & 708 & 2.5e3 & 115.6 & 1.4e4 & 369 & 39.74 & 37.94 & 41.12 & 36.97 & 41.30 \\
            & GPTQ\ts{\S} & 44.01 & 36.77 & 28.14 & - & - & - & - & - & - & - \\
            & AWQ & 2.6e5 & 2.2e5 & 1.2e5 & - & - & - & - & - & - & - \\
            & OmniQuant\ts{\S} & 9.72 & 11.06 & 8.26 & - & - & - & - & - & - & - \\
            & PEQA (our impl.)\ts{\S} & 8.12 & 7.87 & 6.83 & 12.39 & 8.05 & 60.69 & 60.74 & 64.74 & 57.81 & 64.65 \\
            & \gcb{LR-QAT (ours)\ts{\S}} & \gcb{7.86} & \gcb{7.62} & \gcb{6.57} & \gcb{11.09} & \gcb{7.92} & \gcb{61.60} & \gcb{61.70} & \gcb{66.75} & \gcb{60.46} & \gcb{65.27} \\
            \bottomrule
        \end{tabular}
    }  
    \vspace{-.2cm}
\end{table*}
%

%
%

\paragraph{Weight-only quantization}

%
We summarize our results in Table~\ref{tbl:04_weight_only_quant_results}.
As we can see, in almost most cases {\method} outperforms or is on par with prior weight-only quantization methods across various LLM families and quantization settings, including both per-channel and group-wise quantization.
%
In the case of extremely low-bitwidth regime, our method consistently and significantly outperforms related work, across all settings.

In a few cases, especially in case of group-wise quantization our method did not outpeform OmniQuant.
%
However, OmniQuant uses asymmetric quantization which provides extra degrees of freedom compared to symmetric quantization, which are very helpful in the case of low-bit quantization.
In practice, however, symmetric weight quantization yields more efficient inference~\citep{quantization_whitepaper}.
%
Additionally, techniques like OmniQuant and related techniques are orthogonal to our method and can be used as as initialization of {\method}.


\paragraph{Weight-activation quantization}
\begin{table*}[t]
    \setlength{\tabcolsep}{7.2pt}
    \centering
    \caption{\textbf{Weight and activation quantization results for {\llama}-1/2} (denoted `L1'/`L2', respectively). 
    We report WikiText-2 test set perplexity and zero-shot accuracy of 6 tasks. Numbers marked in bold are SOTA. {\ts{\S}}Uses asymmetric weight quantization. {\ts{*}}Uses a maximum sequence length of 1024 for evaluation.}
    \label{tbl:04_weight_act_quant_results}
    \renewcommand{\arraystretch}{1.0}
    \resizebox{0.85\columnwidth}{!}{%
        \begin{tabular}{clccc!{\color{Gray2}\vline}ccc}
            \toprule
            \multirow{2}{*}{\begin{tabular}{c}\# Bits\\(W-A-KV)\end{tabular}} & \multicolumn{1}{c}{\multirow{2}{*}{Method}} & \multicolumn{3}{c}{WikiText-2 perplexity $\da$} & \multicolumn{3}{c}{Avg. zero-shot accuracy $\ua$} \\[0.25em]
              & & L1-7B & L2-7B & L2-13B & L1-7B & L2-7B & L2-13B \\
            \midrule
            FP16 &  & 5.68 & 5.47 & 4.88 & 69.68 & 70.47 & 73.18 \\
            \midrule
            \multirow{5}{*}{\shortstack{4-8-8}} 
            & RTN & 6.88 & 6.17 & 5.23 & 65.83 & 68.55 & 71.54 \\
            & SmoothQuant & 13.7\ts{*} & - & - & 65.17 & - & - \\
            & LLM-QAT & 11.2\ts{*} & - & - & 68.18 & - & - \\
            & PEQA (our impl.) & 5.89 & 5.72 & 5.08 & 68.53 & 69.11 & 72.49 \\
            & \gcb{LR-QAT (ours)} & \gcb{5.85} & \gcb{5.67} & \gcb{5.04} & \gcb{68.58} & \gcb{69.32} & \gcb{73.18} \\
            \midrule
            \multirow{5}{*}{\shortstack{4-8-4}} 
            & RTN & 7.66 & 6.85 & 5.78 & 62.78 & 65.16 & 67.06 \\
            & SmoothQuant & 163.6\ts{*} & - & - & 45.35 & - & - \\
            & LLM-QAT & 11.6\ts{*} & - & - & 64.75 & - & - \\
            & PEQA (our impl.) & 6.15 & 6.03 & 5.27 & 66.54 & 67.56 & 71.35 \\
            & \gcb{LR-QAT (ours)} & \gcb{6.07} & \gcb{5.90} & \gcb{5.24} & \gcb{67.16} & \gcb{69.84} & \gcb{71.65} \\
            \midrule
            \multirow{6}{*}{\shortstack{4-4-4}} 
            & RTN & 17.75 & 18.98 & 11.37 & 49.60 & 51.75 & 55.07 \\
            & SmoothQuant & 25.25 & 83.12 & 35.88 & 38.42 & - & - \\
            & LLM-QAT & - & - & - & 41.27 & - & - \\
            & LLM-QAT + SQ & - & - & - & 46.43 & - & - \\
            & Outlier Suppression+ & - & - & - & 48.43 & - & - \\
            & OmniQuant\ts{\S} & 11.26 & 14.26 & 12.30 & 52.65 & - & - \\
            & PEQA (our impl.) & 8.60 & 8.72 & 7.23 & 58.39 & 57.93 & 62.26 \\
            & \gcb{LR-QAT (ours)} & \gcb{8.47} & \gcb{8.46} & \gcb{7.15} & \gcb{59.00} & \gcb{58.98} & \gcb{62.65} \\
            \bottomrule
        \end{tabular}
    }  
    \vspace{-.2cm}
\end{table*}
We present our results for weight-activation quantization applied to {\llama}-1/2 models in Table~\ref{tbl:04_weight_act_quant_results}.
{\method} demonstrates superior performance compared to all the PTQ and QAT baselines, consistently outperforming them across all model families and the bitwidth settings.
%
In addition to that, as we decrease the activation bitwidths, the improvement in model performance compared to prior work becomes more pronounced.

This indicates {\method}'s versatility, being readily applicable not only to weight-only quantization but also weight-activation quantization, a setting that allows for a very efficient inference using fixed-point arithmetic.
%
%
%
Further, our method can still be combined with most of the related PTQ methods including OmniQuant that shift the difficulty of activation quantization to weight quantization, and will likely lead to even better results.



    \section{Discussion}
\label{sec:discussion}


\paragraph{Limitations}
A core assumption of {\method} is that a low-rank approximation can compensate the introduced quantization noise. While quantization noise follows a random uniform distribution and is theoretically not low-rank, our results and several prior works~\cite{dettmers2024qlora,xu2023qa, kim2024memory} suggest that in an end-to-end training setup, low-rank approaches can effectively compensate quantization noise.
In our work, we demonstrated the effectiveness of {\method} for LLMs up to 13B parameters. It is unclear how it scales to significantly larger LLMs, though we do not see any reason why our findings should not hold beyond 13B models.
We evaluate {\method} as \textit{extended pretraining} technique with several thousands of iterations. It is unclear how it would perform in case it is used \textit{during pretraining} for millions of iterations.



\paragraph{Impact} 
As {\method} improves memory and inference efficiency of LLMs, we expect mainly positive outcomes from our work.
Efficiently deploying LLMs will help with reducing their high power consumption at inference time. It further helps to move inference from the cloud to edge devices which can overcome potential privacy concerns.
In some cases, quantization might lead to biased predictions, see~\cite{hooker2020characterising} for further discussion.





    \section{Conclusions}
\label{sec:conclusions}


In this paper we propose {\method}, a lightweight and memory-efficient QAT algorithm for LLMs which enables training a 7B LLM on a single consumer grade GPU with 24GB of memory.
Inspired by PEFT methods, we introduce a low-rank reparameterization that is aware of the quantization grid. We further reduce the memory requirements by introducing a downcasting operator involving fixed-point or double-packed integers, and applying checkpointing.
In almost all cases, our method outperforms common PTQ approaches and reaches the same model performance as full-model QAT at the fraction of its memory usage.


    \section*{Acknowledgements}
\label{sec:acknowledgements}


We would like to thank Tijmen Blankevoort, Paul Whatmough, Jorn Peters, and Ties van Rozendaal for valuable discussions and support.


    
    {\small
    \bibliography{references}
    \bibliographystyle{plainnat}
    }

\fi  


\ifappendix

    \clearpage
    \newpage



    
    \appendix

    
    \setcounter{table}{0}
    \renewcommand{\thetable}{A\arabic{table}}
    \section{Extended results}
\label{app:extended_results}
%

In this section, we provide additional detailed results.
%
%
%
%
\begin{table*}[htb]
    \setlength{\tabcolsep}{6pt}
    \centering
    \caption{\textbf{LM-eval weight-only quantization results for {\llama}-1 7B}. We report zero-shot accuracy of 6 tasks (higher is better). {\ts{\S}}Uses asymmetric weight quantization.}
    \renewcommand{\arraystretch}{0.9}
    \resizebox{0.94\columnwidth}{!}{%
        \begin{tabular}{clccccccc}
            \toprule
            {\# Bits} & \multicolumn{1}{c}{Method} & BoolQ & PIQA & Winogrande & ARC-e & ARC-c & HellaSwag & {Avg.} \\
            \midrule
            FP16 &  & 75.05 & 79.16 & 70.01 & 72.85 & 44.80 & 76.21 & 69.68 \\
            \midrule
            \multirow{5}{*}{\shortstack{W4\,pc}} & RTN & 73.18 & 78.78 & 69.14 & 71.38 & 44.37 & 74.22 & 68.51 \\
            & GPTQ\ts{\S} & 67.70 & 76.00 & 66.70 & 66.90 & 43.00 & 69.40 & 64.95 \\
            & LSQ (our impl.) & 73.91 & 78.24 & 69.22 & 70.84 & 43.26 & 74.74 & 68.37 \\
            & PEQA (our impl.) & 74.71 & 78.29 & 70.09 & 70.33 & 42.24 & 75.27 & 68.49 \\
            & \gcb{LR-QAT (ours)} & \gcb{74.13} & \gcb{78.29} & \gcb{70.01} & \gcb{71.21} & \gcb{42.41} & \gcb{75.16} & \gcb{68.54} \\
            \midrule
            \multirow{4}{*}{\shortstack{W4\,g128}} & RTN & 74.77 & 78.51 & 70.64 & 71.30 & 43.60 & 74.74 & 68.93 \\
            & LSQ (our impl.) & 75.90 & 79.22 & 70.01 & 71.42 & 43.34 & 75.15 & 69.17 \\
            & PEQA (our impl.) & 75.75 & 79.17 & 70.17 & 70.75 & 43.60 & 75.71 & 69.19 \\
            & \gcb{LR-QAT (ours)} & \gcb{75.29} & \gcb{78.62} & \gcb{69.61} & \gcb{71.59} & \gcb{44.11} & \gcb{75.67} & \gcb{69.15} \\
            \midrule
            \multirow{4}{*}{\shortstack{W3\,pc}} & RTN & 58.93 & 70.40 & 55.72 & 55.01 & 32.17 & 55.75 & 54.66 \\
            & LSQ (our impl.) & 71.35 & 77.97 & 68.82 & 66.33 & 40.10 & 73.14 & 66.29 \\
            & PEQA (our impl.) & 72.69 & 77.15 & 65.90 & 68.27 & 38.91 & 71.60 & 65.75 \\
            & \gcb{LR-QAT (ours)} & \gcb{73.24} & \gcb{78.18} & \gcb{67.40} & \gcb{67.47} & \gcb{40.53} & \gcb{72.77} & \gcb{66.60} \\
            \midrule
            \multirow{4}{*}{\shortstack{W3\,g128}} & RTN & 69.48 & 76.33 & 64.40 & 64.44 & 38.65 & 67.67 & 63.50 \\
            & LSQ (our impl.) & 71.04 & 77.97 & 68.11 & 68.27 & 40.44 & 73.37 & 66.53 \\
            & PEQA (our impl.) & 71.65 & 78.24 & 68.51 & 68.18 & 40.10 & 73.30 & 66.66 \\
            & \gcb{LR-QAT (ours)} & \gcb{72.84} & \gcb{78.02} & \gcb{67.40} & \gcb{68.52} & \gcb{41.04} & \gcb{73.04} & \gcb{66.81} \\
            \midrule
            \multirow{3}{*}{\shortstack{W2\,pc}} & RTN\ts{\S} & 43.24 & 52.61 & 51.62 & 27.48 & 26.28 & 26.26 & 37.92 \\
            & PEQA (our impl.)\ts{\S} & 67.19 & 74.32 & 61.80 & 59.97 & 32.42 & 63.28 & 59.83 \\
            & \gcb{LR-QAT (ours)\ts{\S}} & \gcb{68.07} & \gcb{74.27} & \gcb{64.80} & \gcb{61.03} & \gcb{36.95} & \gcb{65.48} & \gcb{61.77} \\
            \midrule
            \multirow{3}{*}{\shortstack{W2\,g128}} & RTN\ts{\S} & 41.16 & 58.05 & 50.20 & 34.97 & 23.12 & 30.93 & 39.74 \\
            & PEQA (our impl.)\ts{\S} & 67.40 & 74.21 & 62.04 & 60.98 & 35.32 & 64.16 & 60.69 \\
            & \gcb{LR-QAT (ours)\ts{\S}} & \gcb{67.65} & \gcb{74.86} & \gcb{61.80} & \gcb{62.37} & \gcb{37.03} & \gcb{65.90} & \gcb{61.60} \\
            \bottomrule
        \end{tabular}
    }  
    \vspace{-.15cm}
\end{table*}

\vspace{2em}

%
%
\begin{table*}[htb]
    \setlength{\tabcolsep}{6pt}
    \centering
    \caption{\textbf{LM-eval weight-only quantization results for {\llama}-2 7B}. We report zero-shot accuracy of 6 tasks (higher is better). {\ts{\S}}Uses asymmetric weight quantization. }
    \renewcommand{\arraystretch}{0.9}
    \resizebox{0.94\columnwidth}{!}{%
        \begin{tabular}{clccccccc}
            \toprule
            {\# Bits} & \multicolumn{1}{c}{Method} & BoolQ & PIQA & Winogrande & ARC-e & ARC-c & HellaSwag & {Avg.} \\
            \midrule
            FP16 &  & 77.74 & 79.11 & 69.14 & 74.58 & 46.25 & 75.98 & 70.47 \\
            \midrule
            \multirow{4}{*}{\shortstack{W4\,pc}} & RTN & 76.36 & 78.07 & 68.19 & 71.21 & 44.80 & 74.65 & 68.88 \\
            & LSQ (our impl.) & 76.69 & 78.15 & 68.12 & 71.84 & 44.18 & 74.76 & 68.96 \\
            & PEQA (our impl.) & 77.49 & 78.24 & 69.61 & 70.96 & 43.52 & 75.54 & 69.23 \\
            & \gcb{LR-QAT (ours)} & \gcb{77.41} & \gcb{78.56} & \gcb{69.42} & \gcb{72.80} & \gcb{44.66} & \gcb{75.45} & \gcb{69.72} \\
            \midrule
            \multirow{4}{*}{\shortstack{W4\,g128}} & RTN & 76.76 & 78.18 & 69.77 & 72.60 & 45.73 & 75.43 & 69.75 \\
            & LSQ (our impl.) & 77.28 & 78.45 & 69.61 & 72.18 & 44.97 & 75.61 & 69.68 \\
            & PEQA (our impl.) & 76.88 & 78.89 & 69.85 & 72.18 & 44.11 & 75.95 & 69.64 \\
            & \gcb{LR-QAT (ours)} & \gcb{76.73} & \gcb{78.62} & \gcb{70.48} & \gcb{72.85} & \gcb{44.97} & \gcb{75.62} & \gcb{69.88} \\
            \midrule
            \multirow{4}{*}{\shortstack{W3\,pc}} & RTN & 46.27 & 60.28 & 54.85 & 38.05 & 23.29 & 40.47 & 43.87 \\
            & LSQ (our impl.) & 74.39 & 77.91 & 66.85 & 69.15 & 41.53 & 73.04 & 67.14 \\
            & PEQA (our impl.) & 71.62 & 76.82 & 66.14 & 65.66 & 39.76 & 72.63 & 65.44 \\
            & \gcb{LR-QAT (ours)} & \gcb{75.08} & \gcb{77.73} & \gcb{67.50} & \gcb{69.97} & \gcb{42.70} & \gcb{73.24} & \gcb{67.70} \\
            \midrule
            \multirow{4}{*}{\shortstack{W3\,g128}} & RTN & 66.42 & 75.57 & 65.19 & 64.90 & 38.14 & 68.96 & 63.20 \\
            & LSQ (our impl.) & 74.68 & 77.80 & 68.35 & 71.76 & 44.11 & 73.46 & 68.36 \\
            & PEQA (our impl.) & 75.38 & 77.97 & 68.59 & 70.62 & 42.32 & 73.74 & 68.10 \\
            & \gcb{LR-QAT (ours)} & \gcb{76.61} & \gcb{77.31} & \gcb{68.98} & \gcb{72.05} & \gcb{42.58} & \gcb{74.20} & \gcb{68.62} \\
            \midrule
            \multirow{3}{*}{\shortstack{W2\,pc}} & RTN\ts{\S} & 39.57 & 50.71 & 50.20 & 26.73 & 26.28 & 25.63 & 36.52 \\
            & PEQA (our impl.)\ts{\S} & 64.98 & 71.33 & 58.09 & 55.22 & 33.11 & 61.66 & 57.40 \\
            & \gcb{LR-QAT (ours)\ts{\S}} & \gcb{69.88} & \gcb{73.07} & \gcb{62.90} & \gcb{56.57} & \gcb{34.04} & \gcb{63.71} & \gcb{60.03} \\
            \midrule
            \multirow{3}{*}{\shortstack{W2\,g128}} & RTN\ts{\S} & 39.08 & 55.93 & 50.99 & 29.84 & 24.06 & 27.71 & 37.94 \\
            & PEQA (our impl.)\ts{\S} & 69.51 & 73.83 & 63.46 & 58.21 & 34.90 & 64.53 & 60.74 \\
            & \gcb{LR-QAT (ours)\ts{\S}} & \gcb{70.18} & \gcb{73.83} & \gcb{63.69} & \gcb{60.27} & \gcb{34.98} & \gcb{67.24} & \gcb{61.70} \\
            \bottomrule
        \end{tabular}
    }  
    \vspace{-.15cm}
\end{table*}

%
%
\begin{table*}[htb]
    \setlength{\tabcolsep}{6pt}
    \centering
    \caption{\textbf{LM-eval weight-only quantization results for {\llama}-2 13B}. We report zero-shot accuracy of 6 tasks (higher is better). {\ts{\S}}Uses asymmetric weight quantization. {\oom} denotes out of memory.}
    \renewcommand{\arraystretch}{0.9}
    \resizebox{0.95\columnwidth}{!}{%
        \begin{tabular}{clccccccc}
            \toprule
            {\# Bits} & \multicolumn{1}{c}{Method} & BoolQ & PIQA & Winogrande & ARC-e & ARC-c & HellaSwag & {Avg.} \\
            \midrule
            FP16 &  & 80.55 & 80.52 & 72.22 & 77.44 & 48.98 & 79.38 & 73.18 \\
            \midrule
            \multirow{4}{*}{\shortstack{W4\,pc}} & RTN & 79.30 & 79.71 & 70.01 & 75.51 & 48.89 & 76.96 & 71.73 \\
            & LSQ (our impl.) & {\oom} & {\oom} & {\oom} & {\oom} & {\oom} & {\oom} & {\oom} \\
            & PEQA (our impl.) & 78.99 & 80.14 & 71.27 & 76.43 & 48.98 & 79.24 & 72.51 \\
            & \gcb{LR-QAT (ours)} & \gcb{80.15} & \gcb{80.09} & \gcb{72.06} & \gcb{77.65} & \gcb{49.91} & \gcb{79.28} & \gcb{73.19} \\
            \midrule
            \multirow{4}{*}{\shortstack{W4\,g128}} & RTN & 81.10 & 79.82 & 72.38 & 76.73 & 49.06 & 78.52 & 72.94 \\
            & LSQ (our impl.) & {\oom} & {\oom} & {\oom} & {\oom} & {\oom} & {\oom} & {\oom} \\
            & PEQA (our impl.) & 80.28 & 80.63 & 71.74 & 76.14 & 48.38 & 79.62 & 72.80 \\
            & \gcb{LR-QAT (ours)} & \gcb{80.73} & \gcb{80.30} & \gcb{71.74} & \gcb{76.14} & \gcb{49.06} & \gcb{79.51} & \gcb{72.91} \\
            \midrule
            \multirow{4}{*}{\shortstack{W3\,pc}} & RTN & 55.05 & 71.06 & 54.22 & 56.19 & 32.25 & 61.27 & 55.01 \\
            & LSQ (our impl.) & {\oom} & {\oom} & {\oom} & {\oom} & {\oom} & {\oom} & {\oom} \\
            & PEQA (our impl.) & 74.28 & 78.67 & 69.06 & 74.87 & 45.99 & 76.00 & 69.81 \\
            & \gcb{LR-QAT (ours)} & \gcb{78.62} & \gcb{79.49} & \gcb{72.61} & \gcb{73.99} & \gcb{45.56} & \gcb{77.05} & \gcb{71.22} \\
            \midrule
            \multirow{4}{*}{\shortstack{W3\,g128}} & RTN & 74.65 & 76.93 & 69.14 & 70.16 & 42.66 & 72.06 & 67.60 \\
            & LSQ (our impl.) & {\oom} & {\oom} & {\oom} & {\oom} & {\oom} & {\oom} & {\oom} \\
            & PEQA (our impl.) & 78.56 & 78.73 & 69.85 & 73.61 & 44.28 & 76.69 & 70.29 \\
            & \gcb{LR-QAT (ours)} & \gcb{79.79} & \gcb{79.60} & \gcb{70.64} & \gcb{74.24} & \gcb{46.76} & \gcb{78.00} & \gcb{71.51} \\
            \midrule
            \multirow{3}{*}{\shortstack{W2\,pc}} & RTN\ts{\S} & 38.35 & 48.97 & 48.54 & 27.78 & 27.99 & 25.97 & 36.27 \\
            & PEQA (our impl.)\ts{\S} & 71.19 & 74.48 & 59.67 & 62.96 & 37.29 & 68.14 & 62.29 \\
            & \gcb{LR-QAT (ours)\ts{\S}} & \gcb{72.60} & \gcb{76.61} & \gcb{66.22} & \gcb{66.67} & \gcb{39.76} & \gcb{69.79} & \gcb{65.28} \\
            \midrule
            \multirow{3}{*}{\shortstack{W2\,g128}} & RTN\ts{\S} & 50.12 & 57.29 & 50.36 & 34.43 & 22.18 & 32.35 & 41.12 \\
            & PEQA (our impl.)\ts{\S} & 70.34 & 76.88 & 66.85 & 64.94 & 38.74 & 70.66 & 64.74 \\
            & \gcb{LR-QAT (ours)\ts{\S}} & \gcb{75.47} & \gcb{77.86} & \gcb{65.98} & \gcb{67.63} & \gcb{41.04} & \gcb{72.50} & \gcb{66.75} \\
            \bottomrule
        \end{tabular}
    }  
    \vspace{-.15cm}
\end{table*}

%
%
\begin{table*}[htb]
    \setlength{\tabcolsep}{6pt}
    \centering
    \caption{\textbf{LM-eval weight-only quantization results for {\llama}-3 8B}. We report zero-shot accuracy of 6 tasks (higher is better). {\ts{\S}}Uses asymmetric weight quantization. }
    \label{tbl:A_weight_only_zero_shot}
    \renewcommand{\arraystretch}{0.9}
    \resizebox{0.95\columnwidth}{!}{%
        \begin{tabular}{clccccccc}
            \toprule
            {\# Bits} & \multicolumn{1}{c}{Method} & BoolQ & PIQA & Winogrande & ARC-e & ARC-c & HellaSwag & {Avg.} \\
            %
            %
            \midrule
            FP16 &  & 81.44 & 80.79 & 72.85 & 77.74 & 53.33 & 79.16 & 74.22 \\

            \midrule
            \multirow{4}{*}{\shortstack{W4\,pc}} & RTN & 79.02 & 78.56 & 72.85 & 75.97 & 49.32 & 77.44 & 72.19 \\
            & LSQ (our impl.) & 80.58 & 80.03 & 72.69 & 78.24 & 50.17 & 77.94 & 73.28 \\
            & PEQA (our impl.) & 79.57 & 78.67 & 72.93 & 77.19 & 51.11 & 77.25 & 72.79 \\
            & \gcb{LR-QAT (ours)} & \gcb{81.62} & \gcb{79.98} & \gcb{72.85} & \gcb{78.32} & \gcb{52.05} & \gcb{78.19} & \gcb{73.84} \\
            \midrule
            \multirow{4}{*}{\shortstack{W4\,g128}} & RTN & 79.48 & 79.27 & 73.56 & 75.08 & 48.81 & 77.61 & 72.30 \\
            & LSQ (our impl.) & 80.24 & 80.36 & 73.01 & 77.23 & 50.43 & 78.60 & 73.31 \\
            & PEQA (our impl.) & 80.98 & 80.14 & 72.61 & 76.18 & 49.57 & 78.45 & 72.99 \\
            & \gcb{LR-QAT (ours)} & \gcb{80.40} & \gcb{80.90} & \gcb{73.48} & \gcb{77.44} & \gcb{51.11} & \gcb{78.60} & \gcb{73.66} \\
            \midrule
            \multirow{4}{*}{\shortstack{W3\,pc}} & RTN & 58.65 & 61.75 & 56.04 & 39.60 & 23.81 & 44.91 & 47.46 \\
            & LSQ (our impl.) & 76.42 & 78.24 & 70.01 & 72.31 & 45.22 & 75.26 & 69.58 \\
            & PEQA (our impl.) & 63.18 & 64.74 & 57.62 & 43.39 & 26.88 & 50.48 & 51.05 \\
            & \gcb{LR-QAT (ours)} & \gcb{77.46} & \gcb{78.51} & \gcb{69.85} & \gcb{74.83} & \gcb{47.35} & \gcb{74.73} & \gcb{70.46} \\
            \midrule
            \multirow{4}{*}{\shortstack{W3\,g128}} & RTN & 65.47 & 68.39 & 65.19 & 54.00 & 33.45 & 59.96 & 57.74 \\
            & LSQ (our impl.) & 75.50 & 78.78 & 69.69 & 73.57 & 48.55 & 74.55 & 70.11 \\
            & PEQA (our impl.) & 72.26 & 76.06 & 67.80 & 69.02 & 46.08 & 71.89 & 67.19 \\
            & \gcb{LR-QAT (ours)} & \gcb{72.97} & \gcb{79.38} & \gcb{71.67} & \gcb{74.37} & \gcb{49.06} & \gcb{75.44} & \gcb{70.48} \\
            \midrule
            \multirow{3}{*}{\shortstack{W2\,pc}} & RTN\ts{\S} & 44.89 & 48.97 & 47.51 & 25.29 & 27.73 & 26.41 & 36.80 \\
            & PEQA (our impl.)\ts{\S} & 62.54 & 72.96 & 59.51 & 56.82 & 33.70 & 60.77 & 57.72 \\
            & \gcb{LR-QAT (ours)\ts{\S}} & \gcb{65.81} & \gcb{72.31} & \gcb{64.72} & \gcb{53.54} & \gcb{33.45} & \gcb{61.11} & \gcb{58.49} \\
            \midrule
            \multirow{3}{*}{\shortstack{W2\,g128}} & RTN\ts{\S} & 38.47 & 53.32 & 51.78 & 28.75 & 22.70 & 26.81 & 36.97 \\
            & PEQA (our impl.)\ts{\S} & 58.53 & 72.52 & 61.96 & 57.41 & 35.32 & 61.14 & 57.81 \\
            & \gcb{LR-QAT (ours)\ts{\S}} & \gcb{67.98} & \gcb{73.83} & \gcb{62.83} & \gcb{56.94} & \gcb{36.43} & \gcb{64.77} & \gcb{60.46} \\
            \bottomrule
        \end{tabular}
    }  
    \vspace{-.15cm}
\end{table*}

%
%
\begin{table*}[htb]
    \setlength{\tabcolsep}{6pt}
    \centering
    \caption{\textbf{LM-eval weight-only quantization results for Mistral 7B}. We report zero-shot accuracy of 6 tasks (higher is better). {\ts{\S}}Uses asymmetric weight quantization. }
    \renewcommand{\arraystretch}{0.9}
    \resizebox{0.95\columnwidth}{!}{%
        \begin{tabular}{clccccccc}
            \toprule
            {\# Bits} & \multicolumn{1}{c}{Method} & BoolQ & PIQA & Winogrande & ARC-e & ARC-c & HellaSwag & {Avg.} \\
            %
            %
            \midrule
            FP16 &  & 83.58 & 82.10 & 73.88 & 79.59 & 53.92 & 81.07 & 75.69 \\
            \midrule
            \multirow{4}{*}{\shortstack{W4\,pc}} & RTN & 81.22 & 80.63 & 72.53 & 76.77 & 50.09 & 79.41 & 73.44 \\
            & LSQ (our impl.) & 81.96 & 80.41 & 73.09 & 75.08 & 48.55 & 78.19 & 72.88 \\
            & PEQA (our impl.) & 81.80 & 81.12 & 72.61 & 77.23 & 50.17 & 79.43 & 73.73 \\
            & \gcb{LR-QAT (ours)} & \gcb{81.99} & \gcb{81.28} & \gcb{73.56} & \gcb{78.20} & \gcb{51.02} & \gcb{80.57} & \gcb{74.44} \\
            \midrule
            \multirow{4}{*}{\shortstack{W4\,g128}} & RTN & 84.16 & 81.77 & 74.43 & 77.95 & 51.71 & 80.42 & 75.07 \\
            & LSQ (our impl.) & 80.40 & 80.63 & 73.16 & 76.26 & 48.72 & 78.24 & 72.90 \\
            & PEQA (our impl.) & 80.89 & 81.72 & 73.80 & 75.42 & 48.46 & 79.76 & 73.34 \\
            & \gcb{LR-QAT (ours)} & \gcb{83.55} & \gcb{81.61} & \gcb{74.51} & \gcb{78.28} & \gcb{52.90} & \gcb{80.84} & \gcb{75.28} \\
            \midrule
            \multirow{4}{*}{\shortstack{W3\,pc}} & RTN & 68.13 & 77.64 & 63.93 & 63.93 & 41.13 & 72.73 & 64.58 \\
            & LSQ (our impl.) & 79.97 & 80.74 & 70.24 & 74.37 & 47.18 & 77.14 & 71.61 \\
            & PEQA (our impl.) & 80.03 & 80.09 & 69.93 & 72.90 & 45.82 & 77.32 & 71.02 \\
            & \gcb{LR-QAT (ours)} & \gcb{81.62} & \gcb{80.09} & \gcb{70.96} & \gcb{74.75} & \gcb{46.08} & \gcb{77.71} & \gcb{71.87} \\
            \midrule
            \multirow{4}{*}{\shortstack{W3\,g128}} & RTN & 78.44 & 79.60 & 69.14 & 71.17 & 43.00 & 74.75 & 69.35 \\
            & LSQ (our impl.) & 81.01 & 80.36 & 71.03 & 74.41 & 47.27 & 77.68 & 71.96 \\
            & PEQA (our impl.) & 81.99 & 81.18 & 69.61 & 74.92 & 47.18 & 78.37 & 72.21 \\
            & \gcb{LR-QAT (ours)} & \gcb{81.71} & \gcb{80.90} & \gcb{70.48} & \gcb{75.08} & \gcb{47.78} & \gcb{78.50} & \gcb{72.41} \\
            \midrule
            \multirow{3}{*}{\shortstack{W2\,pc}} & RTN\ts{\S} & 39.42 & 51.03 & 49.49 & 26.22 & 26.71 & 26.67 & 36.59 \\
            & PEQA (our impl.)\ts{\S} & 74.34 & 75.95 & 66.22 & 62.21 & 38.48 & 69.68 & 64.48 \\
            & \gcb{LR-QAT (ours)\ts{\S}} & \gcb{75.08} & \gcb{76.39} & \gcb{65.98} & \gcb{64.48} & \gcb{39.59} & \gcb{69.14} & \gcb{65.11} \\
            \midrule
            \multirow{3}{*}{\shortstack{W2\,g128}} & RTN\ts{\S} & 54.98 & 56.91 & 51.70 & 31.27 & 23.21 & 29.70 & 41.30 \\
            & PEQA (our impl.)\ts{\S} & 75.41 & 75.84 & 64.96 & 65.45 & 38.23 & 68.00 & 64.65 \\
            & \gcb{LR-QAT (ours)\ts{\S}} & \gcb{75.81} & \gcb{77.15} & \gcb{64.64} & \gcb{63.59} & \gcb{39.51} & \gcb{70.89} & \gcb{65.27} \\
            \bottomrule
        \end{tabular}
    }  
    \vspace{-.15cm}
\end{table*}

\begin{table*}[htb]
    \setlength{\tabcolsep}{3pt}
    \centering
    \caption{\textbf{LM-eval weight and activation quantization results for {\llama} models}. We report zero-shot accuracy of 6 tasks (higher is better). {\ts{\S}}Uses asymmetric weight quantization. }
    \label{tbl:A_weight_acts_zero_shot}
    \renewcommand{\arraystretch}{1}
    \resizebox{1\columnwidth}{!}{%
        \begin{tabular}{lclccccccc}
            \toprule
            \multicolumn{1}{c}{\multirow{2}{*}{Model}} & \multirow{2}{*}{\begin{tabular}{c}\# Bits\\(W-A-KV)\end{tabular}} & \multicolumn{1}{c}{\multirow{2}{*}{Method}} & \multirow{2}{*}{BoolQ} & \multirow{2}{*}{PIQA} & \multirow{2}{*}{Winogrande} & \multirow{2}{*}{ARC-e} & \multirow{2}{*}{ARC-c} & \multirow{2}{*}{HellaSwag} & \multirow{2}{*}{Avg.} \\
            \\
            \cmidrule[0.5pt]{1-10}
            \multirow{18}{*}{{{\llama}-1 7B}}
            & FP16 &  & 75.05 & 79.16 & 70.01 & 72.85 & 44.80 & 76.21 & 69.68 \\
            \cmidrule{2-10}
            & \multirow{5}{*}{4-8-8} & RTN & 71.35 & 76.66 & 66.46 & 66.84 & 41.55 & 72.10 & 65.83 \\
            & & SmoothQuant & 71.00 & 76.00 & 66.00 & 67.40 & 42.80 & 67.80 & 65.17 \\
            & & LLM-QAT & 74.60 & 77.50 & 67.70 & 70.20 & 45.60 & 73.50 & 68.18 \\
            & & PEQA (our impl.) & 74.86 & 78.24 & 70.01 & 70.12 & 42.83 & 75.14 & 68.53 \\
            & & \gcb{LR-QAT (ours)} & \gcb{73.76} & \gcb{78.51} & \gcb{71.19} & \gcb{71.09} & \gcb{41.81} & \gcb{75.10} & \gcb{68.58} \\
            \cmidrule{2-10}
            & \multirow{5}{*}{4-8-4} & RTN & 68.81 & 75.46 & 62.12 & 62.46 & 39.51 & 68.33 & 62.78 \\
            & & SmoothQuant & 54.70 & 55.40 & 51.50 & 43.90 & 27.70 & 38.90 & 45.35 \\
            & & LLM-QAT & 69.50 & 75.40 & 64.60 & 66.00 & 43.80 & 69.20 & 64.75 \\
            & & PEQA (our impl.) & 72.97 & 77.80 & 67.72 & 67.13 & 40.27 & 73.35 & 66.54 \\
            & & \gcb{LR-QAT (ours)} & \gcb{73.64} & \gcb{77.91} & \gcb{67.56} & \gcb{69.28} & \gcb{41.30} & \gcb{73.25} & \gcb{67.16} \\
            \cmidrule{2-10}
            & \multirow{8}{*}{4-4-4} & RTN & 50.49 & 64.25 & 52.41 & 48.27 & 30.12 & 52.04 & 49.60 \\
            & & SmoothQuant & 49.10 & 49.80 & 48.00 & 30.40 & 25.80 & 27.40 & 38.42 \\
            & & LLM-QAT & 61.30 & 51.50 & 51.90 & 27.90 & 23.90 & 31.10 & 41.27 \\
            & & LLM-QAT + SQ & 62.40 & 55.90 & 50.60 & 35.50 & 26.40 & 47.80 & 46.43 \\
            & & Outlier Suppression+ & 60.21 & 62.73 & 52.96 & 39.98 & 30.29 & 44.39 & 48.43 \\
            & & OmniQuant\ts{\S} & 63.51 & 66.15 & 53.43 & 45.20 & 31.14 & 56.44 & 52.65 \\
            & & PEQA (our impl.) & 65.69 & 72.31 & 59.83 & 56.52 & 34.22 & 61.79 & 58.39 \\
            & & \gcb{LR-QAT (ours)} & \gcb{67.16} & \gcb{71.76} & \gcb{59.59} & \gcb{58.42} & \gcb{34.73} & \gcb{62.34} & \gcb{59.00} \\
            \midrule
            \multirow{10}{*}{{{\llama}-2 7B}}
            & FP16 &  & 77.74 & 79.11 & 69.14 & 74.58 & 46.25 & 75.98 & 70.47 \\
            \cmidrule{2-10}
            & \multirow{3}{*}{4-8-8} & RTN & 75.87 & 77.91 & 67.88 & 71.09 & 44.03 & 74.51 & 68.55 \\
            & & PEQA (our impl.) & 77.37 & 77.97 & 69.77 & 70.54 & 43.52 & 75.50 & 69.11 \\
            & & \gcb{LR-QAT (ours)} & \gcb{77.00} & \gcb{78.13} & \gcb{69.14} & \gcb{72.10} & \gcb{44.11} & \gcb{75.44} & \gcb{69.32} \\
            \cmidrule{2-10}
            & \multirow{3}{*}{4-8-4} & RTN & 70.37 & 76.01 & 63.38 & 68.94 & 41.47 & 70.76 & 65.16 \\
            & & PEQA (our impl.) & 74.71 & 77.48 & 67.40 & 69.28 & 42.75 & 73.75 & 67.56 \\
            & & \gcb{LR-QAT (ours)} & \gcb{74.46} & \gcb{77.69} & \gcb{68.51} & \gcb{69.78} & \gcb{42.75} & \gcb{73.82} & \gcb{67.84} \\
            \cmidrule{2-10}
            & \multirow{3}{*}{4-4-4} & RTN & 57.86 & 64.91 & 54.46 & 49.62 & 31.83 & 51.83 & 51.75 \\
            & & PEQA (our impl.) & 67.09 & 70.67 & 60.06 & 54.80 & 32.17 & 62.76 & 57.93 \\
            & & \gcb{LR-QAT (ours)} & \gcb{66.94} & \gcb{71.98} & \gcb{60.77} & \gcb{57.20} & \gcb{33.87} & \gcb{63.10} & \gcb{58.98} \\ 
            \midrule
            \multirow{10}{*}{{{\llama}-2 13B}}
            & FP16 &  & 80.55 & 80.52 & 72.22 & 77.44 & 48.98 & 79.38 & 73.18 \\
            \cmidrule{2-10}
            & \multirow{3}{*}{4-8-8} & RTN & 79.24 & 79.27 & 70.01 & 75.51 & 48.29 & 76.92 & 71.54 \\
            & & PEQA (our impl.) & 79.02 & 80.20 & 71.19 & 76.60 & 48.72 & 79.22 & 72.49 \\
            & & \gcb{LR-QAT (ours)} & \gcb{81.10} & \gcb{79.76} & \gcb{71.35} & \gcb{77.36} & \gcb{50.60} & \gcb{78.91} & \gcb{73.18} \\
            \cmidrule{2-10}
            & \multirow{3}{*}{4-8-4} & RTN & 77.25 & 76.61 & 66.69 & 67.72 & 41.13 & 72.98 & 67.06 \\
            & & PEQA (our impl.) & 78.01 & 79.22 & 69.30 & 75.59 & 48.21 & 77.78 & 71.35 \\
            & & \gcb{LR-QAT (ours)} & \gcb{78.59} & \gcb{79.54} & \gcb{70.80} & \gcb{75.29} & \gcb{48.04} & \gcb{77.64} & \gcb{71.65} \\
            \cmidrule{2-10}
            & \multirow{3}{*}{4-4-4} & RTN & 62.60 & 67.90 & 53.20 & 57.32 & 34.13 & 55.25 & 55.07 \\
            & & PEQA (our impl.) & 68.72 & 74.21 & 62.35 & 63.85 & 36.01 & 68.44 & 62.26 \\
            & & \gcb{LR-QAT (ours)} & \gcb{70.64} & \gcb{73.88} & \gcb{63.14} & \gcb{61.78} & \gcb{38.14} & \gcb{68.31} & \gcb{62.65} \\
            \bottomrule
        \end{tabular}
    }  
    \vspace{-.1cm}
\end{table*}

\begin{table*}[htb]
    \setlength{\tabcolsep}{6pt}
    \centering
    \caption{\textbf{A comparison between min-max and the best range setting used for round-to-nearest (RTN) initialization for {\llama} and Mistral models}. 
    We report WikiText-2 test set perplexity (lower is better) and average zero-shot accuracy (higher is better). {\ts{\S}}Uses asymmetric weight quantization.}
    \label{tbl:A_rtn}
    \renewcommand{\arraystretch}{1.0}
    \resizebox{1.0\columnwidth}{!}{%
        \begin{tabular}{clccccc!{\color{Gray2}\vline}ccccc}
            \toprule
            \multirow{2}{*}{\# Bits} & \multicolumn{1}{c}{Range} & \multicolumn{5}{c}{WikiText-2 perplexity $\da$} & \multicolumn{5}{c}{Avg. zero-shot accuracy $\ua$} \\[0.1em]
              & \multicolumn{1}{c}{estimator} & L1-7B & L2-7B & L2-13B & L3-8B & M-7B & L1-7B & L2-7B & L2-13B & L3-8B & M-7B \\
            \midrule
            FP16 &  & 5.68 & 5.47 & 4.88 & 6.14 & 5.25 & 69.68 & 70.47 & 73.18 & 74.22 & 75.69 \\
            \midrule
            \multirow{3}{*}{\shortstack{W4\,pc}} 
            & best est. & $L^4$ & $L^{3.5}$ & $L^{3.5}$ & $L^{3.5}$ & $L^4$ & $L^4$ & $L^{3.5}$ & $L^{3.5}$ & $L^{3.5}$ & $L^4$ \\
            & best & 6.33 & 6.14 & 5.21 & 7.53 & 5.91 & 68.51 & 68.88 & 71.73 & 72.19 & 73.44 \\
            & min-max & 6.85 & 7.14 & 5.40 & 10.53 & 6.33 & 66.23 & 66.41 & 72.19 & 67.44 & 71.84 \\
            \midrule
            \multirow{3}{*}{\shortstack{W4\,g128}} 
            & best est. & $L^5$ & min-max & min-max & $L^4$ & $L^5$ & $L^5$ & min-max & min-max & $L^4$ & $L^5$ \\
            & best & 6.05 & 5.78 & 5.04 & 6.96 & 5.49 & 68.93 & 69.75 & 72.94 & 72.30 & 75.07 \\
            & min-max & 6.08 & 5.78 & 5.04 & 6.99 & 5.51 & 68.96 & 69.75 & 72.94 & 72.95 & 74.98 \\
            \midrule    
            \multirow{3}{*}{\shortstack{W3\,pc}} 
            & best est. & $L^{3.5}$ & $L^{3.5}$ & $L^5$ & $L^{3.5}$ & $L^4$ & $L^{3.5}$ & $L^{3.5}$ & $L^5$ & $L^{3.5}$ & $L^4$ \\
            & best & 12.88 & 26.73 & 8.71 & 34.10 & 9.49 & 54.66 & 43.87 & 55.01 & 47.46 & 64.58 \\
            & min-max & 2.4e4 & 1.9e4 & 2.3e3 & 1.6e5 & 3.2e3 & 36.02 & 35.71 & 37.85 & 35.78 & 36.78 \\
            \midrule
            \multirow{3}{*}{\shortstack{W3\,g128}} 
            & best est. & $L^5$ & $L^4$ & $L^5$ & $L^5$ & $L^5$ & $L^5$ & $L^4$ & $L^5$ & $L^5$ & $L^5$ \\
            & best & 7.95 & 7.61 & 6.20 & 15.11 & 6.77 & 63.50 & 63.20 & 67.60 & 57.74 & 69.35 \\
            & min-max & 8.10 & 8.22 & 6.14 & 29.38 & 7.22 & 62.69 & 64.07 & 66.81 & 54.54 & 68.35 \\
            \midrule
            \multirow{3}{*}{\shortstack{W2\,pc\ts{\S}}} 
            & best est. & $L^{2.4}$ & $L^3$ & $L^3$ & $L^{3.5}$ & $L^{3.5}$ & $L^{2.4}$ & $L^3$ & $L^3$ & $L^{3.5}$ & $L^{3.5}$ \\
            & best & 4.9e3 & 5.2e3 & 5.2e3 & 6.4e4 & 6.8e3 & 37.92 & 36.52 & 36.27 & 36.80 & 36.59 \\
            & min-max & 1.1e5 & 2.5e4 & 4.9e4 & 1.4e6 & 7.5e4 & 39.07 & 36.53 & 39.01 & 37.69 & 38.13 \\
            \midrule
            \multirow{3}{*}{\shortstack{W2\,g128\ts{\S}}} 
            & best est. & $L^4$ & $L^{3.5}$ & $L^5$ & $L^4$ & $L^5$ & $L^4$ & $L^{3.5}$ & $L^5$ & $L^4$ & $L^5$ \\
            & best & 708 & 2.5e3 & 115.6 & 1.4e4 & 369 & 39.74 & 37.94 & 41.12 & 36.97 & 41.30 \\
            & min-max & 3.6e3 & 5.9e3 & 341 & 2.8e5 & 3.4e3 & 37.06 & 36.78 & 40.30 & 37.18 & 37.16 \\
            \bottomrule
        \end{tabular}
    }  
    \vspace{-.2cm}
\end{table*}
\begin{table*}[htb]
    \setlength{\tabcolsep}{6pt}
    \centering
    \caption{\textbf{Runtime comparison} between LR-QAT, full-model QAT (LSQ), full-precision training and full-precision LoRA for {\llama} 7B on Nvidia A100 80GB GPU, assuming effective batch size 32 and sequence length 1024. We repeat each experiment 5 times and report mean $\pm{}$ standard deviation. {\oom} denotes out of memory.
    }
    \label{tbl:A_runtime}
    \resizebox{0.95\columnwidth}{!}{%
    \begin{tabular}{ llcc }
        \toprule
        \multirow{2}{*}{\# Bits} & \multicolumn{1}{c}{\multirow{2}{*}{Method}} & {Per-device batch size $\times{}$}  & \multicolumn{1}{c}{\multirow{2}{*}{Time/100 steps, sec}} \\
        & & grad. accumulation steps & \\
        \midrule
        FP16 & Full-model training & 1$\times{}$32 & \ms{974}{3} \\
        FP16 & LoRA, $r=32$ & 1$\times{}$32 & \ms{950}{3} \\
        \midrule
        W4\,pc & Full-model QAT & 1$\times{}$32 & {\oom} \\
        W4\,pc & Full-model QAT + CPU opt. offloading & 1$\times{}$32 & \ms{7426}{216} \\
        W4\,pc & Full-model QAT + CPU opt. offloading & 2$\times{}$16 & {\oom} \\
        W4\,pc & Full-model QAT + checkpointing & 1$\times{}$32 & \ms{3248}{7} \\
        W4\,pc & Full-model QAT + checkpointing & 2$\times{}$16 & {\oom} \\
        \midrule
        W4\,pc & {\method}, $\dcfn{} = {\text{\nf{Q4.4}}}$, $r=32$ & 1$\times{}$32 & \ms{2938}{6} \\
        W4\,pc & {\method}, $\dcfn{} = {\text{\nf{Q4.4}}}$, $r=32$ & 4$\times{}$8 & \ms{1522}{5} \\
        \midrule
        W4\,pc & {\method}, $\dcfn{} = {\text{\nf{Q4.4}}}$, $r=1$ & 4$\times{}$8 & \ms{1519}{6} \\
        W4\,pc & {\method}, $\dcfn{} = {\text{\nf{Q4.4}}}$, $r=4$ & 4$\times{}$8 & \ms{1528}{3} \\
        W4\,pc & {\method}, $\dcfn{} = {\text{\nf{Q4.4}}}$, $r=256$ & 4$\times{}$8 & \ms{1546}{4} \\
        \midrule
        W4\,pc & {\method}, $\dcfn{} = {\text{\nf{INT4}}}$, $r=32$ & 4$\times{}$8 & \ms{1518}{8} \\
        W4\,g128 & {\method}, $\dcfn{} = {\text{\nf{Q4.4}}}$, $r=32$ & 4$\times{}$8 & \ms{1528}{5} \\
        \bottomrule
    \end{tabular}
    }  
    \vspace{-.15cm}
\end{table*}
%

%


    %
    \setcounter{table}{0}
    \renewcommand{\thetable}{B\arabic{table}}
    \clearpage
\section{Experimental details}
\label{app:exp_details}
%

%
%
In this section, we list the details related to hyperparameters and other settings used in our experiments.
If not stated otherwise, the standard hyperparameters that we use are the one shown in Table~\ref{tbl:B_training_hyperparams}.
\begin{table*}[htb]
    \centering
    \begin{tabular}{ ll }
        \toprule
         Hyperparameter & Value / Search space \\
         \midrule  
         Optimizer & {AdamW} \\
         Learning rate for $\A$, $\B$ ({\nf{INT4}}\,/\,{\nf{INT3}}\,/\,{\nf{INT2}}) & $\{10^{-5}, 10^{-4}, 10^{-3}, 10^{-2}\}$ \\
         Learning rate for $\s$ ({\nf{INT4}}\,/\,{\nf{INT3}}) & $\{0{\ts{*}}, 10^{-5}\}$ \\
         Learning rate for $\s$ ({\nf{INT2}}) & $\{10^{-5}, 10^{-4}\}$ \\
         Learning rate for $\z$ ({\nf{INT2}} only) & $0{\ts{*}}$ \\
         Learning rate for $\W$ (full-model QAT only) & $\{1,5,10,50,100\}\cdot10^{-5}$ \\
         Learning rate schedule for $\A$, $\B$ & \texttt{linear} (with warmup) \\
         Learning rate schedule for $\s$ & \texttt{linear} (with warmup) \\
         Learning rate schedule for $\W$ (full-model QAT only) & \texttt{linear} (with warmup) \\
         Weight decay for $\A$, $\B$ & $0$ \\
         Weight decay for $\s$ & $0$ \\
         Weight decay for $\W$ (full-model QAT only) & $0.1$ \\
         Adam $\p{\beta_1, \beta_2}$ & $\p{0.9, 0.95}$ \\
         Training steps & $10^4$ \\
         Warmup steps & $10\%$ of {Training steps} \\
         Batch size & $32$ \\
         Maximum sequence length (during training) & $1024$ \\
         $L^2$-norm gradient clipping (maximum norm) & $1.0$ \\
         $\alpha$ in~\eqref{eq:lr_qat} & $1.0$ \\
         \bottomrule
    \end{tabular}
    \vspace{-.1cm}
    \caption{
    Common hyperparameters used for experiments.
    {\ts{*}}Is equivalent to freezing the quantization scale / zero offset obtained after initial range estimation ($\s = \sO$, $\z = \zO$).
}
    \label{tbl:B_training_hyperparams}
\end{table*}

%
 

\paragraph{Quantization}
We experiment with both weight-only and weight-activation quantization. 
The default settings are {\nf{INT4}}\,/\,{\nf{INT3}}\,/\,{\nf{INT2}} per-channel (denoted `pc') and group-wise weight quantization with a group size of 128 (denoted `g128').
We use symmetric quantization, except the {\nf{INT2}} case, where we use asymmetric quantization~\eqref{eq:dequant}, for a fair comparison with related work.
We quantize all linear layers, except the classification head.
In weight-activation quantization, defaults are {\nf{INT4}} per-channel weight and
per-token activation quantization~\citep{dettmersgpt3}.
Following OmniQuant~\citep{shao2023omniquant}, we quantize all inputs to matmuls with exception of the softmax output and additionally quantize the KV-cache as in LLM-QAT~\citep{liu2023llm}.
%

\paragraph{Libraries}
We implement our method in PyTorch~\citep{pytorch} and use training and evaluation pipelines from HuggingFace libraries~\cite{accelerate,lhoest-etal-2021-datasets,wolf2020transformers}.
For zero-shot evaluation, we use the LM Evaluation Harness framework~\citep{gao2021framework}.
Specifically, we use lm\_eval v0.4.2 and report~\texttt{acc\_norm} for tasks where it's available (PIQA, ARC-e, ARC-c, HellaSwag) and otherwise~\texttt{acc} (BoolQ and Winogrande).

\paragraph{Datasets and training}
To optimize the learnable parameters, we use AdamW optimizer~\citep{loshchilov2017decoupled} with weight decay set to zero, $(\beta_1, \beta_2) = (0.9, 0.95)$ and linear learning rate warm up over the first $10\%$ steps, following by a linear decay to zero by the end of training.
We use a separate maximum learning rate for quantization scales and for low-rank adapters, which are tuned depending on the experiment.

We apply our methods to all linear layers in the attention blocks (both in self-attention and in the feed-forward network).
We only train low-rank auxiliary matrices $\A$, $\B$ and the quantization parameters $\s$.
Specifically, we freeze embedding layers, the final classification heads and RMSNorm parameters.
In the case of asymmetric weight quantization, a zero offset $\z$ is set during range estimation phase and kept frozen throughout training.

We train on a small subset of SlimPajama~\citep{slimpajama}, which is a close open-source replica of the dataset used for pre-training {\llama} models.
%
%
We select hyperparameters based on the perplexity of a small subset of Wikipedia validation set\footnote{Specifically, we use the English subset of Wiki40b,~\url{https://huggingface.co/datasets/wiki40b}, that contains cleaned-up text of English Wikipedia and training/validation splits.} (512 sequences).
%
For all weight-only and weight-activation quantization results, including the comparison with full-model QAT in Section~\ref{subsec:comparison_with_full_model_qat}, we train for $10^4$ steps. 
%
For ablation studies in Sections~\ref{subsec:ablation_rank} and~\ref{subsec:ablation_phi_init_dtype} we use shorter training of $10^3$ steps. 
Since the full-model QAT experiment requires more than 80GB of GPU memory, we apply checkpointing on the quantization function $\What$ to be able to run the experiment on an Nvidia A100 80GB GPU.
We have also experimented with CPU optimizer state offloading, but that turned out to be significantly slower, see details in Table~\ref{tbl:A_runtime}.
%
%
%
%
The rest of the hyperparameters and their search spaces are listed in Table~\ref{tbl:B_training_hyperparams}.

\paragraph{PTQ initialization}
%
We compare with vanilla round-to-nearest quantization (RTN), where we explore several choices of range setting and report the best one based on Wikipedia validation set perplexity, and also use that as initialization for our method.
Specifically, we experimented with min-max range estimator and with $L^p$-norm range estimator with the following values for $p$: $\{2.0, 2.4, 3.0, 3.5, 4.0, 5.0\}$.


\paragraph{Computational Resources}
All the experiments were executed on a single Nvidia A100 GPU equipped with 80GB of VRAM.
Models of size 7B, 8B and 13B needed respectively around 2.25, 2.75 and 4.5 days for $10^4$ training steps experiments.
For obtaining all the results in the paper, including the ablations, we needed 202 GPU days (A100). 
Including preliminary experiments that did not make it in the final paper and hyperparameter turning we estimate the total compute costs of this research to approximately 750 GPU days.

\fi  

\end{document}